\documentclass[letterpaper, 10 pt, conference]{ieeeconf}  

\IEEEoverridecommandlockouts                              

\title{\LARGE \bf
	Collision Recovery Control of a Foldable Quadrotor
}

\author{Karishma Patnaik, Shatadal Mishra, Zachary Chase, and Wenlong Zhang$^{*}$
	\thanks{The authors are with The Polytechnic School, Ira A. Fulton Schools of Engineering, Arizona State University, Mesa, AZ, 85212, USA. Email: {\tt\small $\{$kpatnaik, smishr13, ztchase, wenlong.zhang$\}$@asu.edu}.}%
	\thanks{$^{*}$Address all correspondence to this author.}%
}
\usepackage{url}
\usepackage{amsmath}
\usepackage{tikz}
\usepackage{textcomp}
\usepackage{lipsum}
\usepackage{graphicx}
\usepackage{amssymb}
\usepackage{mathtools}
\usepackage{xcolor}
\usepackage{float}
\usepackage{epsfig}
\usepackage[outdir=./]{epstopdf}
\usepackage[backend=bibtex,sorting=none]{biblatex}

\usepackage[caption=false]{subfig}
\bibliography{bibliography}
\usepackage[bottom=1.1in, top=52pt,right=46pt,left=46pt]{geometry}
\usepackage[caption=false]{subfig}
\usepackage[pdfencoding=auto, psdextra]{hyperref}

\newcommand\copyrighttext{%
	\footnotesize \textcopyright 2021 IEEE. Personal use of this material is permitted.
	Permission from IEEE must be obtained for all other uses, in any current or future 
	media, including reprinting/republishing this material for advertising or promotional 
	purposes, creating new collective works, for resale or redistribution to servers or 
	lists, or reuse of any copyrighted component of this work in other works. 
}
\newcommand\copyrightnotice{%
	\begin{tikzpicture}[remember picture,overlay]
		\node[anchor=south,yshift=10pt] at (current page.south) {\fbox{\parbox{\dimexpr\textwidth-\fboxsep-\fboxrule\relax}{\copyrighttext}}};
	\end{tikzpicture}%
}

\begin{document}
	
	\maketitle
	\copyrightnotice
	\thispagestyle{empty}
	\pagestyle{empty}
	
	\begin{abstract}
		
		Autonomous missions of small unmanned aerial vehicles (UAVs) are prone to collisions owing to environmental disturbances and localization errors. Consequently, a UAV that can endure collisions and perform recovery control in critical aerial missions is desirable to prevent loss of the vehicle and/or payload. We address this problem by proposing a novel foldable quadrotor system which can sustain collisions and recover safely. The quadrotor is designed with integrated mechanical compliance using a torsional spring such that the impact time is increased and the net impact force on the main body is decreased. The post-collision dynamics is analysed and a recovery controller is proposed which stabilizes the system to a hovering location without additional collisions. Flight test results on the proposed and a conventional quadrotor demonstrate that for the former, integrated spring-damper characteristics reduce the rebound velocity and lead to simple recovery control algorithms in the event of unintended collisions as compared to a rigid quadrotor of the same dimension. 
	\end{abstract}

	\section{Introduction}
	Quadrotor systems are found in a multitude of applications like aerial photography, package delivery, surveillance, search and rescue missions, and aerial manipulation tasks \cite{OS12,ML18}. In such missions, a quadrotor is often required to fly through cluttered environments. Although obstacle avoidance methods are employed to ensure flight safety, a quadrotor system is still susceptible to collisions due to unknown environmental disturbances and/or imperfect sensor data. This warrants the need for a collision-resilient feature in quadrotor systems which stabilizes the system after collision. In this context, mechanical robustness coupled with recovery control is necessary to complete missions successfully and reliably \cite{P+20}.
	\par 
	During high-impact collisions, all the energy is directly transferred to the chassis because of minimal stopping distance \cite{C01, DZ95}. Since the chassis carries bulk of the payload, it is imperative to protect it from high impacts for achieving safe missions. Some commercially available quadrotor platforms use foam rings in the propeller guard to absorb collision energy, which is effective only at low-energy collisions due to low specific stiffness \cite{BC11}. Another common solution to withstand collisions is to encase the system in a protective structure \cite{BC09, KB13}. Apart from the significant increase in weight and decrease in payload capacity, encasing the complete system in a protective structure will disable many applications such as object manipulation.  
	\par 
	Besides design modifications, post-collision recovery control is also crucial to develop successful collision-resilient quadrotors. For this reason, one of the main requirements is to detect collisions instantaneously and accurately. Briod et al. \cite{BP13} use contact sensors and onboard accelerometers to exploit contacts with the environment and autonomously navigate like insects. These techniques require a robust mechanical frame that is also optimized for sensor placement to recognize collisions in any direction \cite{BS12, NL13}. As an alternative, researchers are exploring external wrench estimation methods to accurately determine collision for subsequent interaction control \cite{OM15}. Ruggiero et al. \cite{RC15} propose a wrench estimator using a momentum-based approach that requires translational velocity information. Naldi et al. \cite{NT13} propose a control loop supervisor which detects collision based on an error threshold of the path-following controller. Tomi{\'c} and Haddadin \cite{TH14} propose a hybrid external wrench estimation technique that combines the momentum-based and acceleration-based approaches to detect collisions and achieve safe reaction. Chui et al. \cite{CD16} and Dicker et al. \cite{DC18} investigate the collision characteristics using a fuzzy logic process and design a control strategy which allows automatic recovery from a non-destructive collision. However, these estimation algorithms are viable at low-collision velocities with small impact forces. 
	\par 
	
	\begin{figure}[t]
		\centering
		\includegraphics[width=0.48\textwidth]{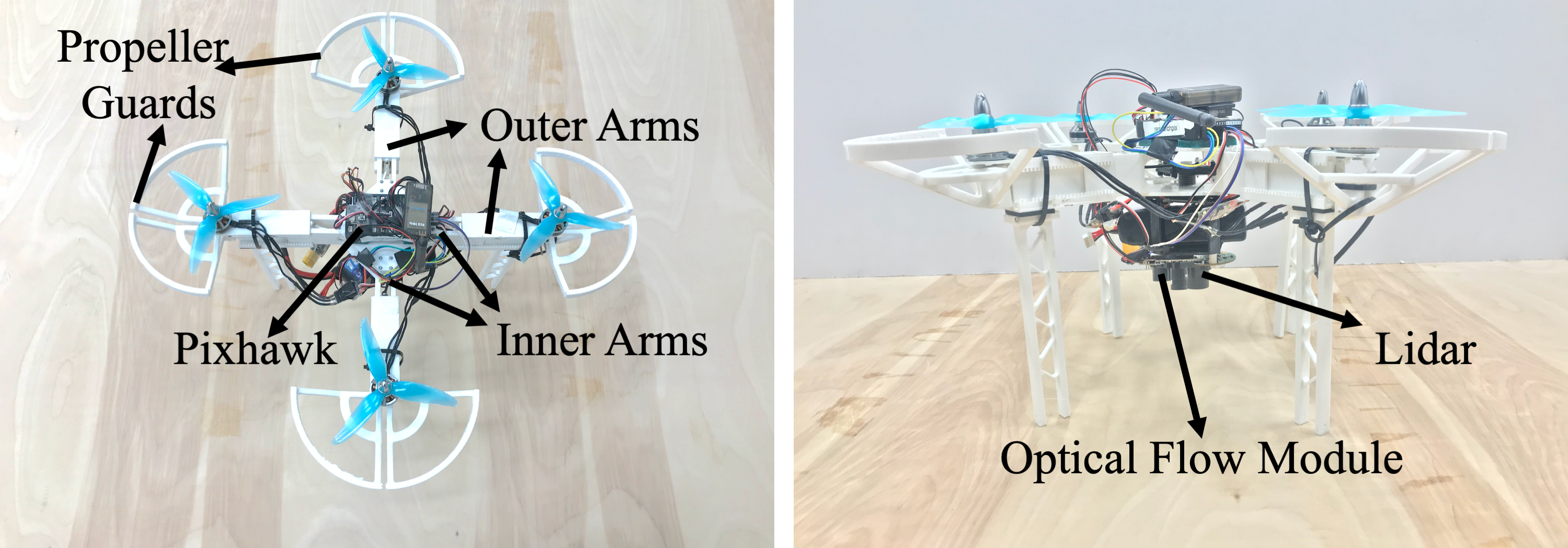}
		\caption{The foldable quadrotor proposed in this paper. The left figure shows the top view and the right figure shows the side view of the quadrotor. A video for the experiments can be found at \url{https://youtu.be/obepryCvdZ4}}
		\label{fig:rigid_foldable}
		\vspace{-0.2in}
	\end{figure}

	\par
	In this paper, we combine design modifications with an admittance based recovery control algorithm to develop a novel collision resilient quadrotor (Fig. \ref{fig:rigid_foldable}) based on passive spring actuation. With the proposed design, we aim to increase the total impact time in the event of unseen collisions by allowing the chassis to fold inwards and admit the collision impact force. This phenomena consequently decreases the impact sustained by the main body and protects it from breaking due to high impact forces in comparison to a conventional quadrotor. A simple and effective recovery control algorithm is proposed such that it stabilizes the vehicle after collision and aids online replanning.
	\par 
	The remainder of this paper is organized as follows:
	Section II details the design and fabrication of the quadrotor for collision recovery. Section III describes dynamic models of the foldable quadrotor developed from physical laws and proposes a novel method to generate the post-collision recovery setpoint. Section IV discusses the simulation results for the proposed method. Section V presents the experimental results. Section VI concludes this paper and discusses future work.
	\section{Hardware}\label{sec:hardware}
	\begin{figure}[t]
		\includegraphics[clip, trim=5cm 5.5cm 4.5cm 3cm, width =0.45\textwidth]{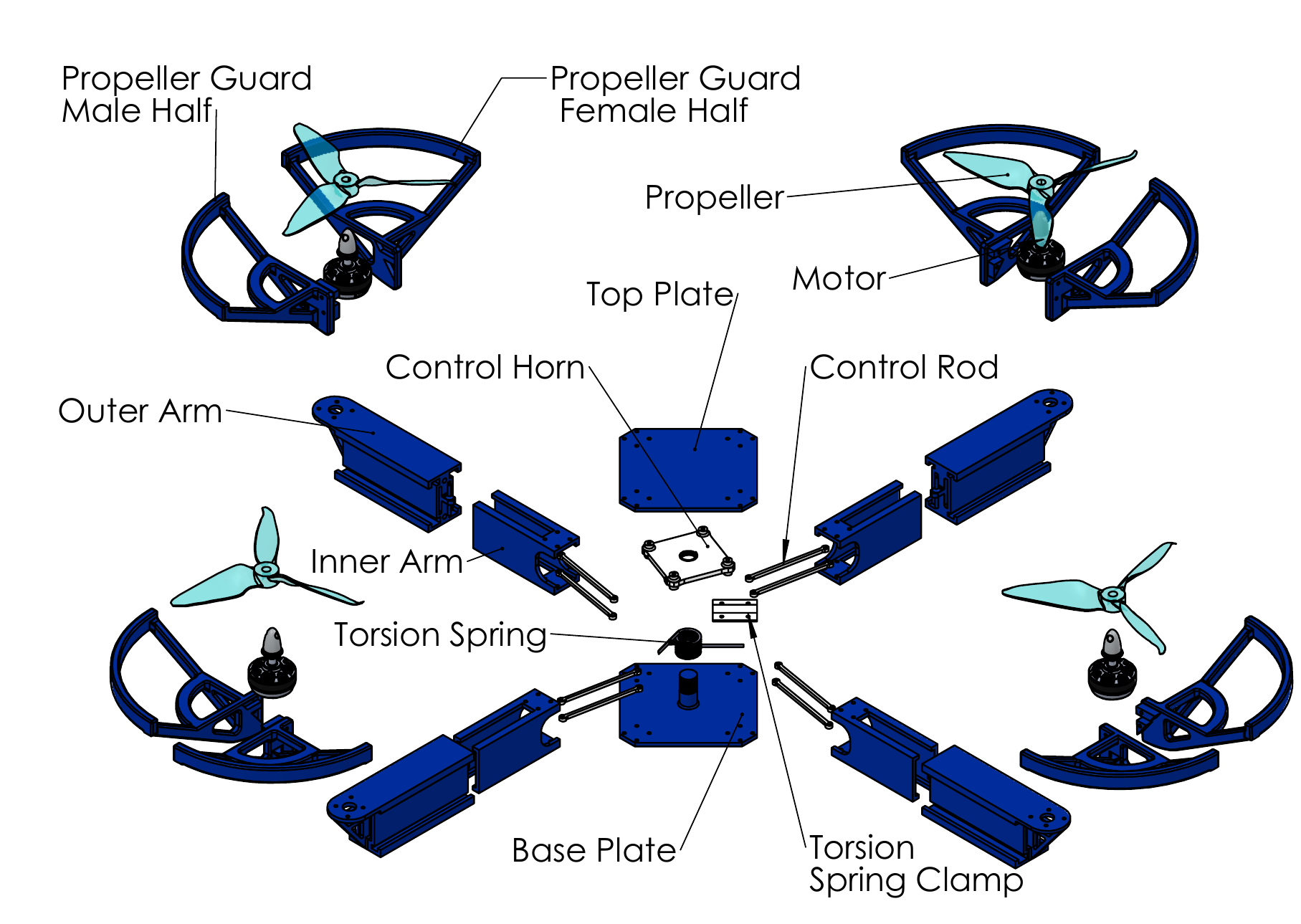}
		\caption{Exploded view of the components of the folding mechanism based on a passive spring}
		\label{fig:design}
		\vspace{-0.2in}
	\end{figure}
\begin{figure}[t]
		\centering
		\subfloat[{Top view of the arms in a retracted position} \label{fig:extend}]
		{\includegraphics[width =0.2\textwidth]{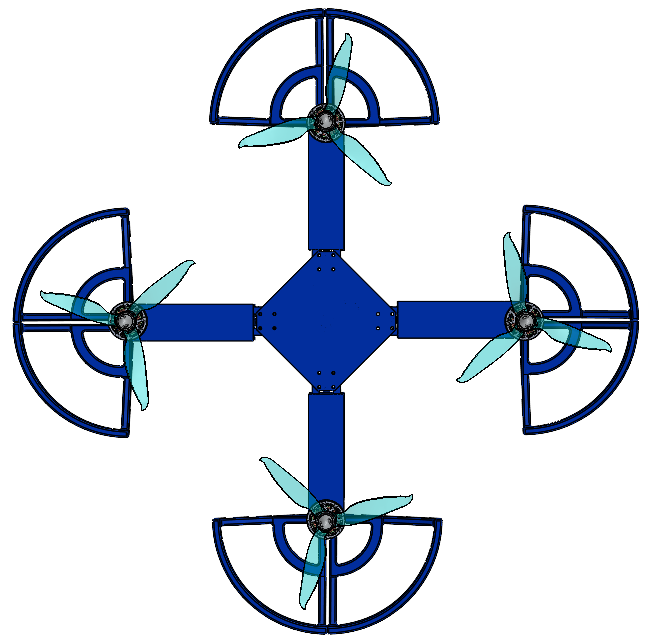}
		}
		\subfloat[{Top view of the arms fully \newline extended out} \label{fig:cont}]
		{\includegraphics[width =0.235\textwidth]{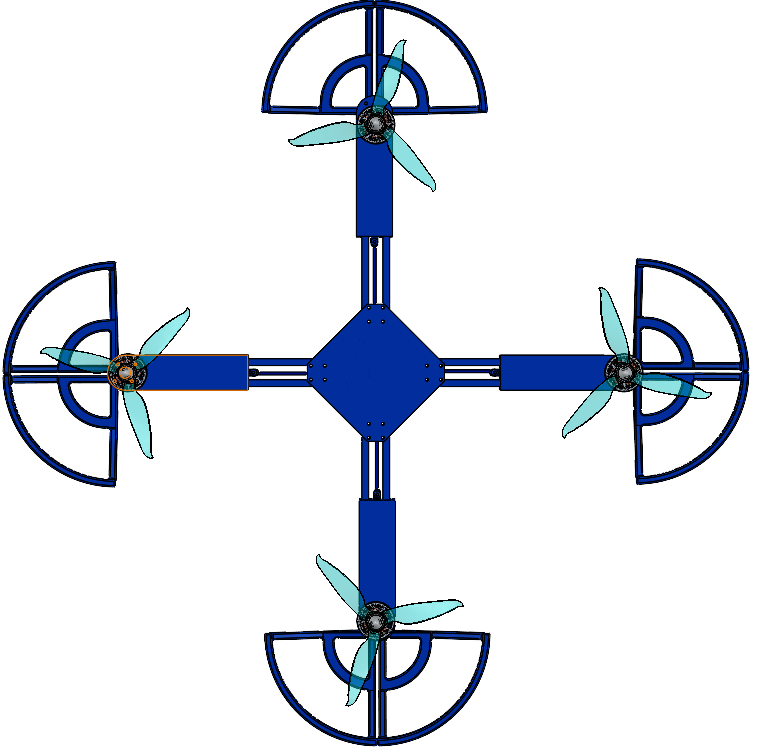}
		}
		\caption{A top view of the quadrotor dimensions after complete retraction during collision. Besides being collision resilient, the fully extended arms have a motor-motor distance of about 290mm and the completely retracted position is around 230mm, a 20\% reduction in the motor-motor distance.}
		\label{fig:topviews}
		\vspace{-0.2in}
	\end{figure}

	The proposed foldable quadrotor is 3D printed and we use polylactide (PLA) for fabricating a lightweight and durable structure. The components of the design are shown in the exploded view of the full assembly in Fig. \ref{fig:design}. The inner arms are connected directly to the base and top plates which act as the anchor points for the arm assembly. The outer arms are allowed to move inward and outward 
	to make the quadrotor foldable upon impact. Two torsion control rods are attached to the end of each inner arm with the other end attached to the control horn. The control horn is threaded onto the base plate which allows the arms to be stable during flight and also allow a slight vertical movement of collapsing during impact. The torsion spring is attached to the bottom of the control horn and to the top of the base plate via the torsion spring clamp. Having two connection points ensures that the torsion spring stays in place and torques during impact, creating the desired collapsing effect during collision. As the quadrotor moves away from the object, the torsion spring returns to its relaxed position pushing the arms back to their original flight position. Fig. \ref{fig:topviews} shows the spring loaded and unloaded configurations of the platform respectively to demonstrate folding upon physical interaction. Finally, the propeller guards act as the point of contact for the arms and are attached in line with the horizontal plane of the arm. This directs majority of the collision force towards the center of mass to reduce the chance of flipping the entire quadrotor body during collision, while also protecting the propeller blades from damage.
	\subsubsection*{Spring Specifications and Structure Weight}
	Easily replaceable, off-the-shelf torsional springs are used for building the quadrotor. The torsional spring should have a spring constant of $0.3Nm/rad$ and return the arms back to their initial positions without breaking or damaging any part of the chassis. This specific torque rating is needed to move the arms back to their original positions based upon the mass of the arms. Additionally, the maximum rotation angle for the spring is constrained by the maximum mechanically feasible angle that the control horn can move. Considering these specifications, a $5.7g$ steel torsion spring with an allowable deflection angle of $120^o$ and a torque value of $0.513Nm$ is selected. The current design provides an inward arm movement of $3cm$ each. The entire mechanical chassis combined with propeller guards weighs approximately $400g$ which is $80g$ heavier than a conventional quadrotor of similar dimensions. The design doesn't produce vibrations due to the appropriate spring chosen and does not hinder the overall performance of the quadrotor compared to a conventional one.
	\begin{figure}[b]
		\centering
		\subfloat[{Cross section} \label{fig:section_view}]
		{\includegraphics[clip, trim=4cm 4cm 4cm 4cm, width =0.12\textwidth]{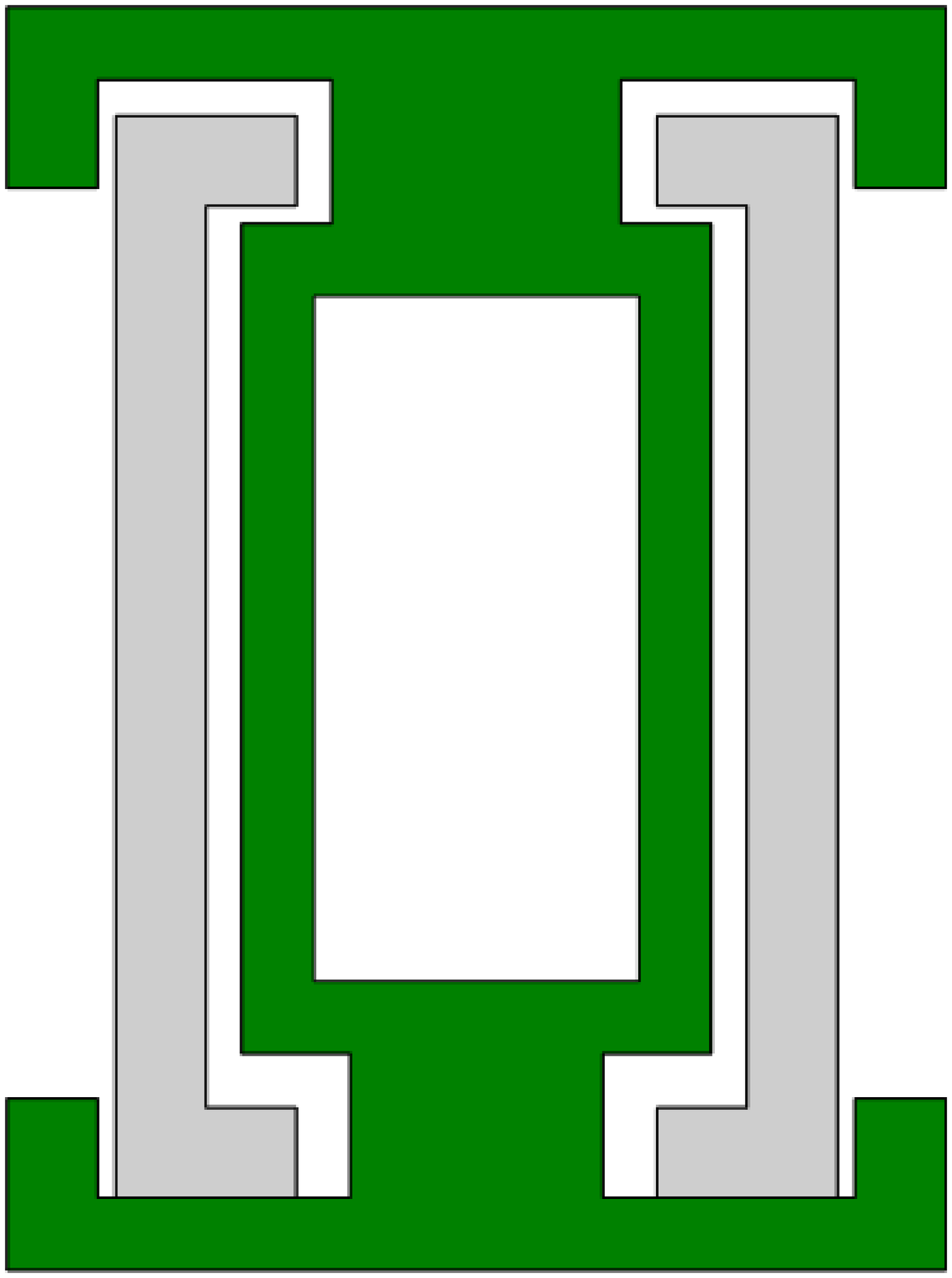}
		}~~~~~
		\subfloat[{Factor of safety plot} \label{fig:fos}]
		{\includegraphics[clip, trim=4cm 16.7cm 2cm 4cm, width =0.3\textwidth]{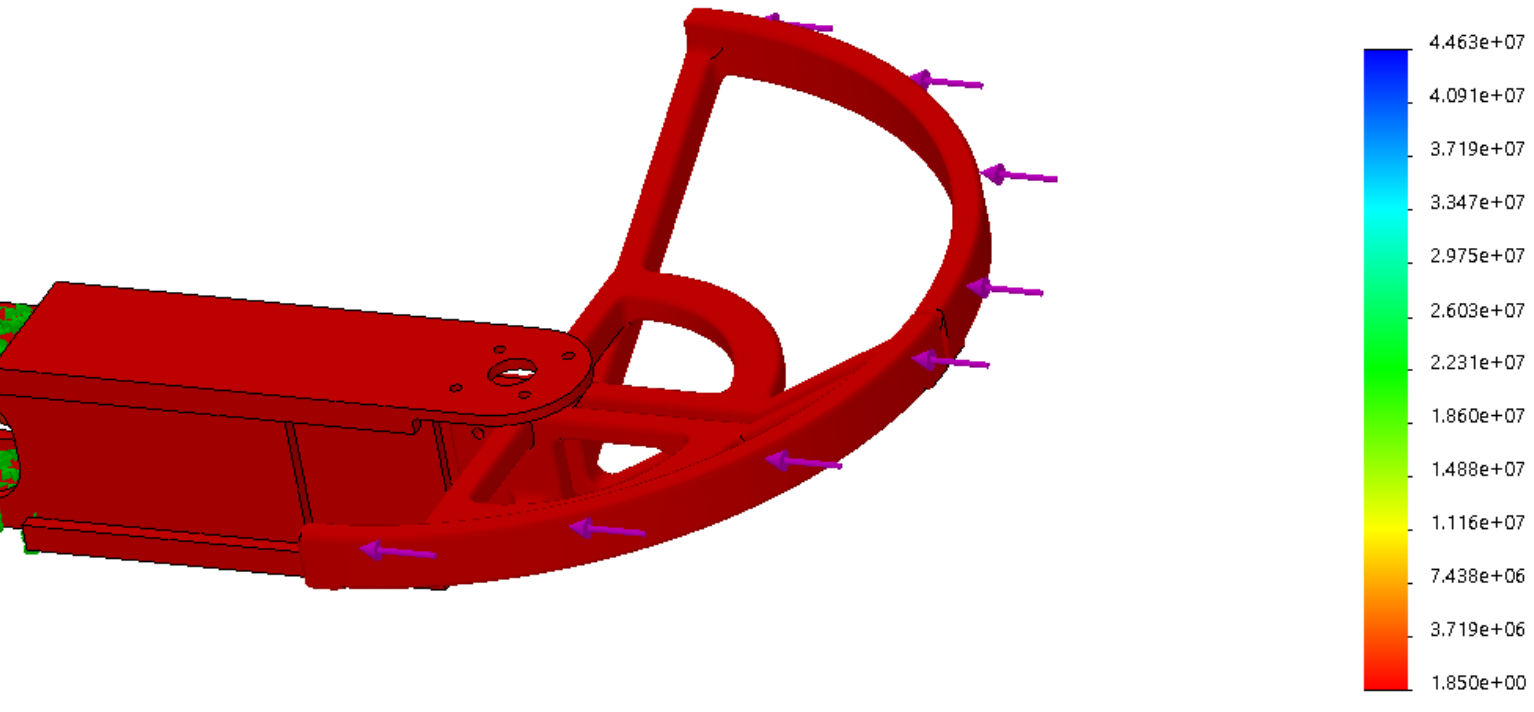}
		}
		\caption{Cross section cut of the arm geometry and factor of safety plot denoting impact force intensity.}
		\label{fig:section}
	\end{figure} 
	\vspace{-0.1in}
	\subsection{Finite Element Analysis}
	According to the chosen motor-propeller combination, the completely assembled quadrotor could weigh a maximum of $1kg$ to keep the hover thrust below 70\%. This will allow the quadrotor to achieve a horizontal velocity of $5m/s$ and above for testing and evaluating the viability of our collision recovery quadrotor at high speeds. 
	The design also has a square channel with cut outs to reduce vibrations during flight as shown in Fig. \ref{fig:section_view}.
	To ensure that the quadrotor could handle impact forces, a finite element analysis (FEA) is performed using SolidWorks 2019. The maximum impact force on the arm for the quadrotor weighing $1kg$ is calculated to be $100N$ for a velocity change of $5m/s$ in $0.05s$. A static FEA is then applied on the single arm to determine the factor-of-safety (FOS) plot and analyze the durability of the modeled components. A remote load was implemented to simulate the quadrotor arm receiving the external load from the propeller guards. Ideally, an FOS value at $2.00$ or higher is the standard value of safety implemented in industry today for most designs and from Fig. \ref{fig:fos}, the minimum FOS of the current design is calculated as $1.85$.
	\section{Analysis And Control}
	\begin{table}[t]
		\caption{Notations used in this paper}
		\centering 
		\begin{tabular}{c c c c}
			\hline\hline
			Symbol & Description \\[0.5ex]
			\hline
			$\{i_1, ~i_2, ~i_3\} $ & inertial frame \\
			$\{b_1, ~b_2, ~b_3\}$ & body-fixed frame \\
			$x \in \mathbb{R}^3 $ & position of vehicle in inertial frame  \\ 
			$v \in \mathbb{R}^3 $ & velocity of vehicle in inertial frame  \\  
			$R \in \mathsf{SO(3)}$ & rotation matrix \\ 
			$\Omega \in \mathbb{R}^3 $ & angular velocity in body frame \\
			$f \in \mathbb{R} $ & thrust input in $b_3$ direction \\
			$M \in \mathbb{R}^3 $ & moment input in body frame \\
			$J \in \mathbb{R}^{3\times3} $ & moment of inertia matrix\\
			$l \in \mathbb{R} $ & change in arm length \\
			$k_p,k_v,k_R,k_\Omega \in \mathbb{R} $ & gains for inner-loop control \\
			[1ex]
			\hline
			\vspace{-0.3in}
		\end{tabular}
		\label{table:nonlin}
	\end{table}
	\subsection{Quadrotor Modeling}
\begin{figure}[t]
		\centering
		\includegraphics[width=0.25\textwidth]{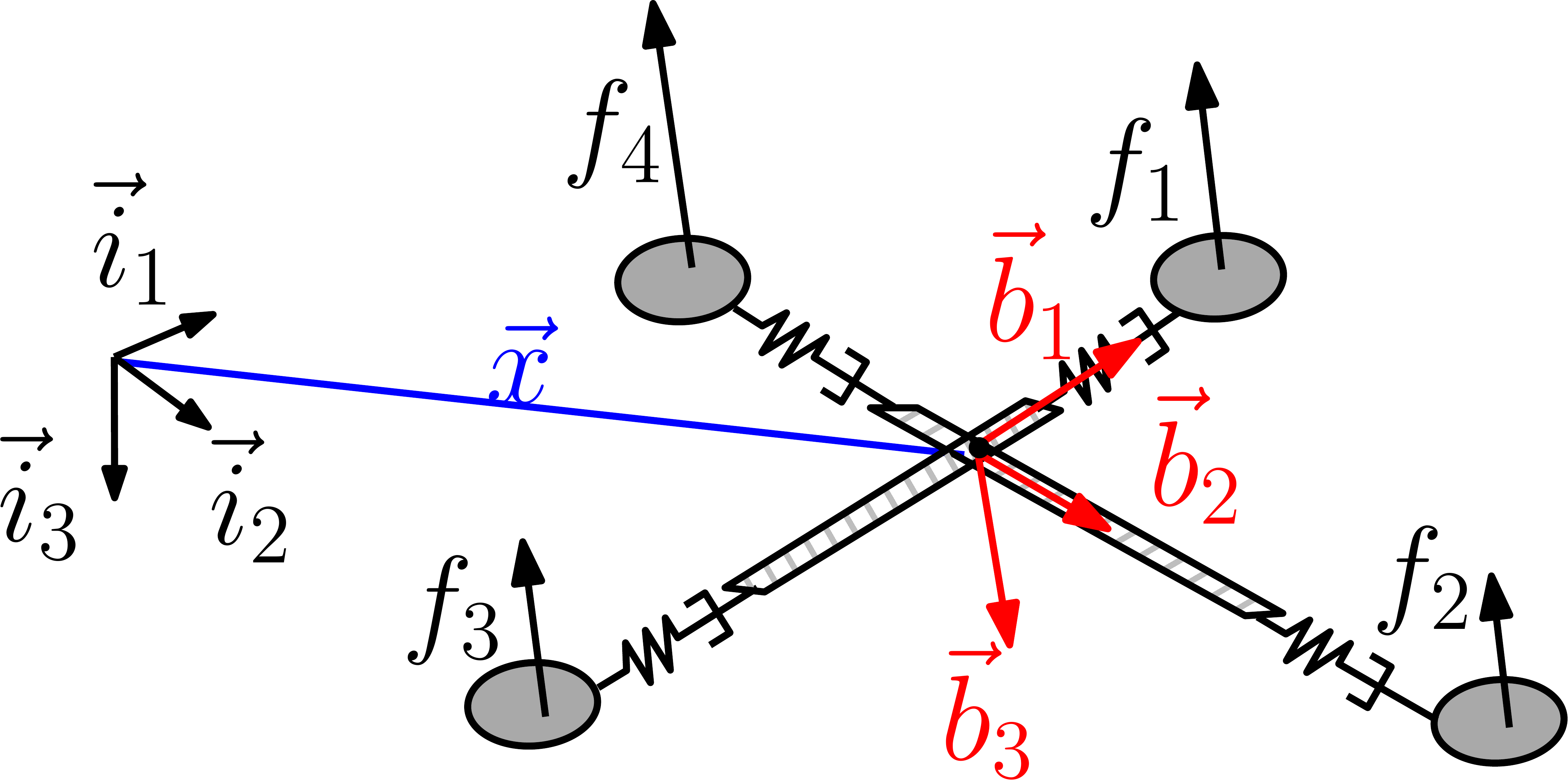}
		\caption{Inertial and body frame description. The quadrotor arms are modelled as spring integrated subsystems with the same spring constant for every arm.}
		\label{fig:axis}
		\vspace{-0.2in}
	\end{figure}

	To model and control the quadrotor system, we choose an inertial frame $ \{i_1, ~i_2, ~i_3\} $ and body-fixed frame $\{b_1, ~b_2, ~b_3\}$ whose origin coincides with the centre-of-gravity of the system as shown in Fig. \ref{fig:axis}. In general, the configuration of a quadrotor at any point of time is given by the coordinates of the center of mass and its orientation in the inertial frame. The translational acceleration of the vehicle is determined by the attitude of the vehicle and the total thrust produced by the four propellers. We assume that the thrust produced by each rotor is individually controlled \cite{LL10}. Therefore, the control inputs are identified as the total thrust, $f \in \mathbb{R}$, of the vehicle and the total moment, $\tau \in \mathbb{R}^3$, in the body fixed frame. The dynamics of the quadrotor can be written as:
	\begin{equation}\label{Eqn:dynamics_S1}
		\begin{aligned}
			m\dot{{v}} &= mg{e}_3 - f{R}{e}_3 \\
			\dot{{x}} &= {v} \\
			\dot{{R}} &= {R}\hat{{\Omega}} \\
			{J}\dot{{\Omega}} &= {\tau} - {\Omega} \times {J}{\Omega} 
		\end{aligned}
	\end{equation}
	where $m$ denotes the vehicle mass, ${x} \in \mathbb{R}^3$ describes the location of center of mass (COM) in the inertial frame, $J$ denotes the moment of inertia, ${v} \in \mathbb{R}^3$ describes the corresponding velocity vector, ${R} \in \mathsf{SO(3)}$ is the rotation matrix from the body-fixed frame to the inertial frame, ${\Omega} \in \mathbb{R}^3 $ denotes the angular velocity vector in the body-fixed frame, $g = 9.81 ~ ms^{-2}$ denotes the acceleration due to gravity, ${e}_3$ denotes the $i_3$ axis unit vector and the \textit{hat map} $\hat{\cdot}: \mathbb{R}^3 \xrightarrow{} \mathrm{so(3)}$ is a symmetric matrix operator defined by the condition that $\hat{x}y = x \times y, ~\forall~ x,y \in \mathbb{R}^3$. 
	\vspace{-0.05in}
	\subsection{Analyzing Post-Collision Dynamics}
	The quadrotor's post-collision dynamics evolve with the initial conditions $[x(t_c), ~v_{rb}]^T$ where $v_{rb}$ is the \textit{rebound velocity} and $t_c$ denotes the time of collision.
	If a new recovery setpoint is not generated at the time of collision, the quadrotor will keep running into the wall mimicking a bouncing ball automaton until it crashes. In order to generate a simple and fast recovery setpoint, we investigate $v_{rb}$ of the foldable quadrotor by analyzing the arm dynamics and exploit this velocity to generate the new setpoint. 
	\subsubsection{Initial Conditions Post Collision}\label{sec:icanalysis}
	During the normal flight, the quadrotor arm maintains constant length, $l_{arm}$. However, upon collision, the torsional spring facilitates the movement of the arm inwards and evokes the spring-damper characteristics of the arm. Therefore, the arm length changes dynamically according to:
	\begin{equation}\label{eqn:spring_soln}
		\vspace{-0.1in}
		\begin{aligned}
			\ddot{l} &= -b_s \dot{l} - k_s l
		\end{aligned}
	\end{equation}
	where $l$ is the change in the arm length, and $b_s$ and $k_s$ are the mass-normalized damping and spring coefficients. Assuming that the impact force is less than or equal to the maximum spring force, the spring dynamics are integrated to find the system velocity when $l = 0$ after the first peak.  Since this is the instant by which the quadrotor exits the wall, this velocity is used as the rebound velocity $v_{rb}$ in the simulations. With this assumption, we proceed to analyze the post collision dynamics using (\ref{Eqn:dynamics_S1}) with initial conditions $[x(t_c), ~v_{rb}]^T$ to simulate tracking performance after collision.
	\subsubsection{Comparison with Rigid Quadrotor}\label{sec:analysis}
	For a rigid quadrotor, the magnitude of $v_{rb} \approx v_{c}$ (collision velocity), due to the high coefficient of restitution from negligible $b_s$ and high $k_s$. However, for the spring integrated arm, impact energy is dissipated in compressing and stabilizing the spring-damper system, consequently leading to low rebound velocities. 
	Therefore, when comparing the tracking performance of a rigid and a foldable system for the \textit{same recovery setpoint} ($x_d$), the latter with significantly low rebound velocities as initial condition, such that $v_{rb}|_{fold} < v_{rb}|_{rigid}$, stabilizes the system at the new setpoint with \textit{no or low} overshoot. However, the rigid system, with a very high rebound velocity as its initial condition, overshoots the desired setpoint. This phenomenon is exploited to generate the new desired setpoint as described in the next section.
	
	\subsection{Proposed 
		Recovery Controller}\label{sec:recoverycontrol}
	Once a collision is detected, we propose a recovery set point in the $i_1-i_2$ plane as a function of the \textit{collision velocity}, $v_c$. We employ $v_c$ to generate an adaptive setpoint for various collision scenarios and not base it on a constant term in order to inherently include the generation of a recovery setpoint farther away if the impact velocities are high. This helps avoid overshoot and oscillatory response in tracking of the new setpoint. Also, the proposed method is easy to implement with low computational expense. 
	
	Let the collision velocity in the $i_1-i_2$ plane be denoted by $v_{c} \in \mathbb{R}^2$, then the post collision recovery setpoint, $x_d$, is updated just once after the collision as:  
	\begin{equation}\label{eqn:rebound}
		x_d(t_c) = \big[x_1(t_c)-\gamma_1 |v_{c_1}|, ~~ x_2(t_c) - \gamma_2 |v_{c_2}|, ~~ x_3(t_c)\big]^T
	\end{equation} 
	where $\gamma_1$ and $\gamma_2$ are positive tuning parameters, $t_c$ denotes the time of collision and the negative sign is used to generate a setpoint in an opposite direction to that of the approach. It is to be noted that in (\ref{eqn:rebound}), the height setpoint, $x_{3d}(t_c)$, does not update and remains the same as the current height. 
	
	We employ an attitude tracking controller on the $\mathsf{SO(3)}$ space to track this setpoint. The errors are defined according to \cite{LL10}: 
	\begin{equation}\label{eqn:errors}
		\begin{aligned}
			e_x &= x_d - x,~ 
			e_v = v_d - v, \\
			e_R &= \frac{1}{2}(R_d^T R - R^T R_d)^\vee, \
			e_\Omega = \Omega - R^T R_d \Omega_d
		\end{aligned}
	\end{equation}
	where the $vee~map ~^\vee : \mathrm{so(3)} \xrightarrow{} \mathbb{R}^3 $ is the inverse of the $hat~map$ and the control moment is generated according to:
	\begin{equation}\label{eqn:controller}
		\begin{aligned}
			\tau = -k_R e_R-k_\Omega e_\Omega +\Omega \times J\Omega + J\alpha \\
		\end{aligned}
	\end{equation}
	where $k_R$ and $k_\Omega$ are appropriate positive constants and $\alpha$ is the angular acceleration term. The position control loop is a cascaded P-PID controller to track position as given below:
	\begin{equation}
		\begin{aligned}
			f &= k_v e_v + k_{v_I} \int e_v dt + k_{v_D}\frac{d}{dt} e_v, \\
			v_d &= k_p e_x
		\end{aligned}
	\end{equation}
	with positive constants $k_v, k_{v_I},k_{v_D}$ and $k_p$.
	\textcolor{black}{The stability proof for the system follows \cite{LL10} but with the inclusion of a dynamically varying $J$ in the attitude control loop. In our previous work \cite{LM19}, we showed the exponential stability of the attitude tracking for a bounded and varying $J$. Due to page limits, we omit the proof in this paper.} 
	
	\begin{figure}[t]
		\centering
		\subfloat[{Parameter identification} \label{fig:coeffs}]
		{\includegraphics[width = 0.23\textwidth]{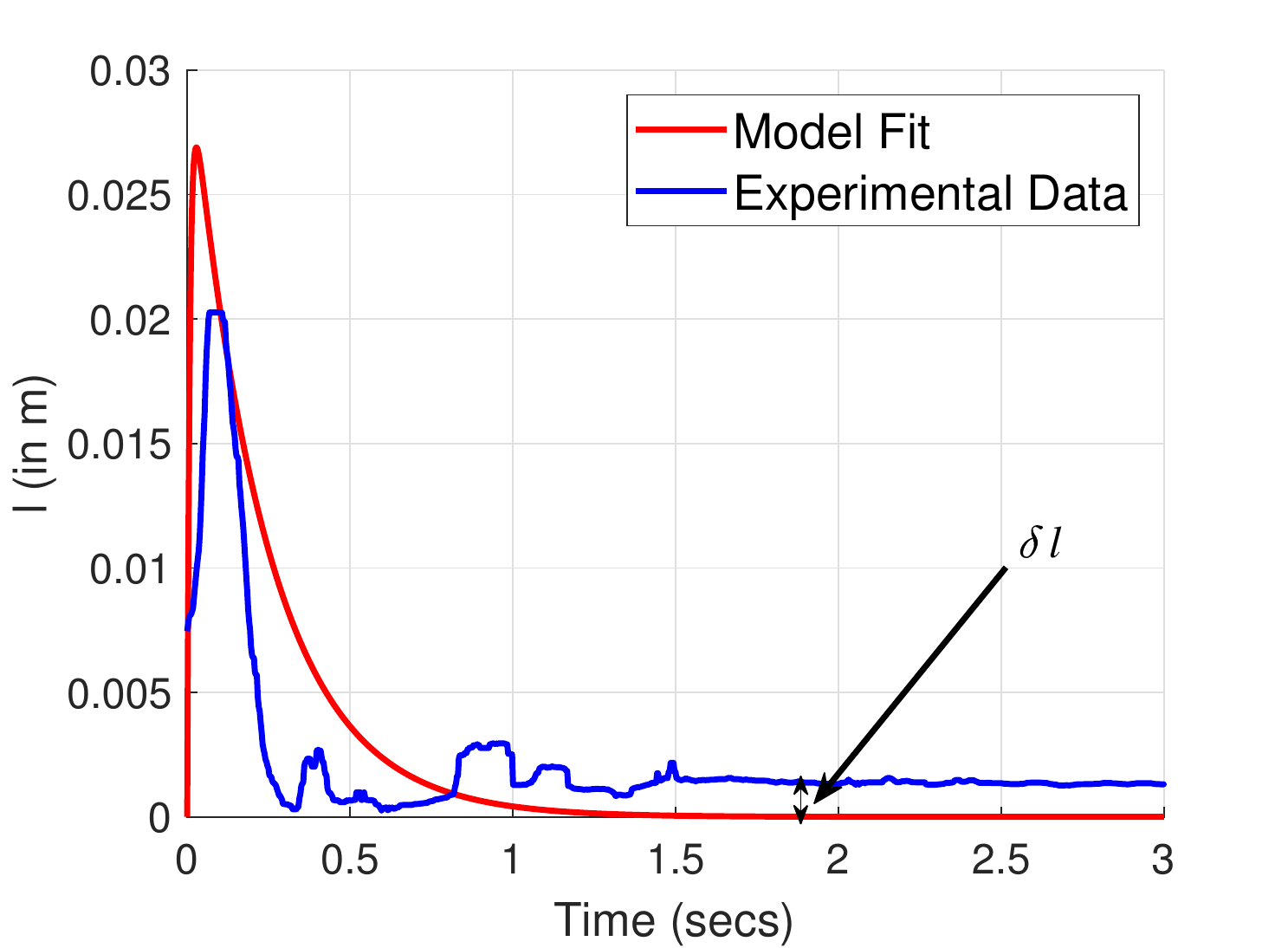}
		}
		\subfloat[{Position trajectory in $i_1$} \label{fig:simx}]
		{\includegraphics[width = 0.23\textwidth]{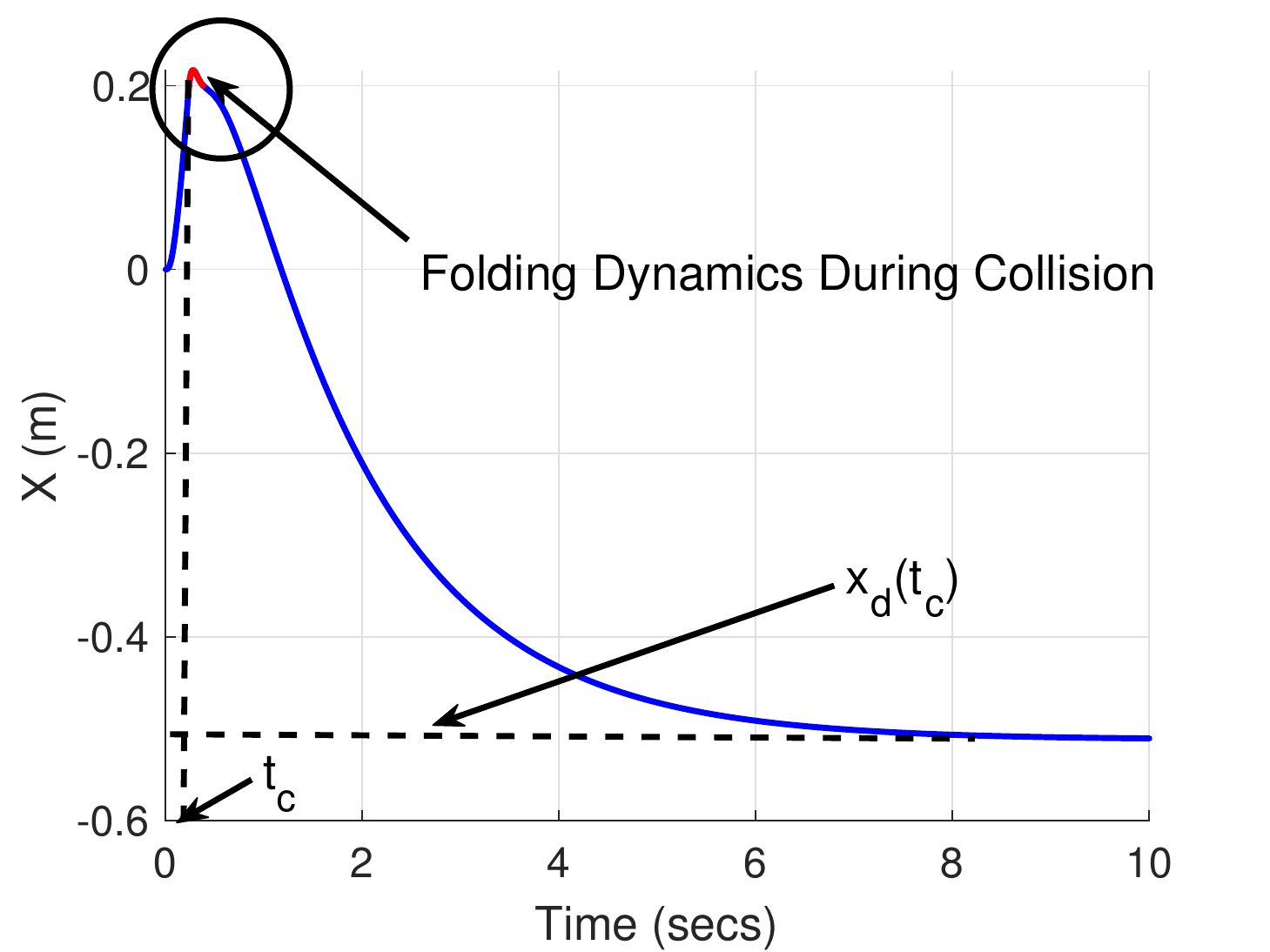}
		}\\
		\subfloat[{ Velocity trajectory in $i_1$ } \label{fig:simexv}]
		{\includegraphics[width = 0.23\textwidth]{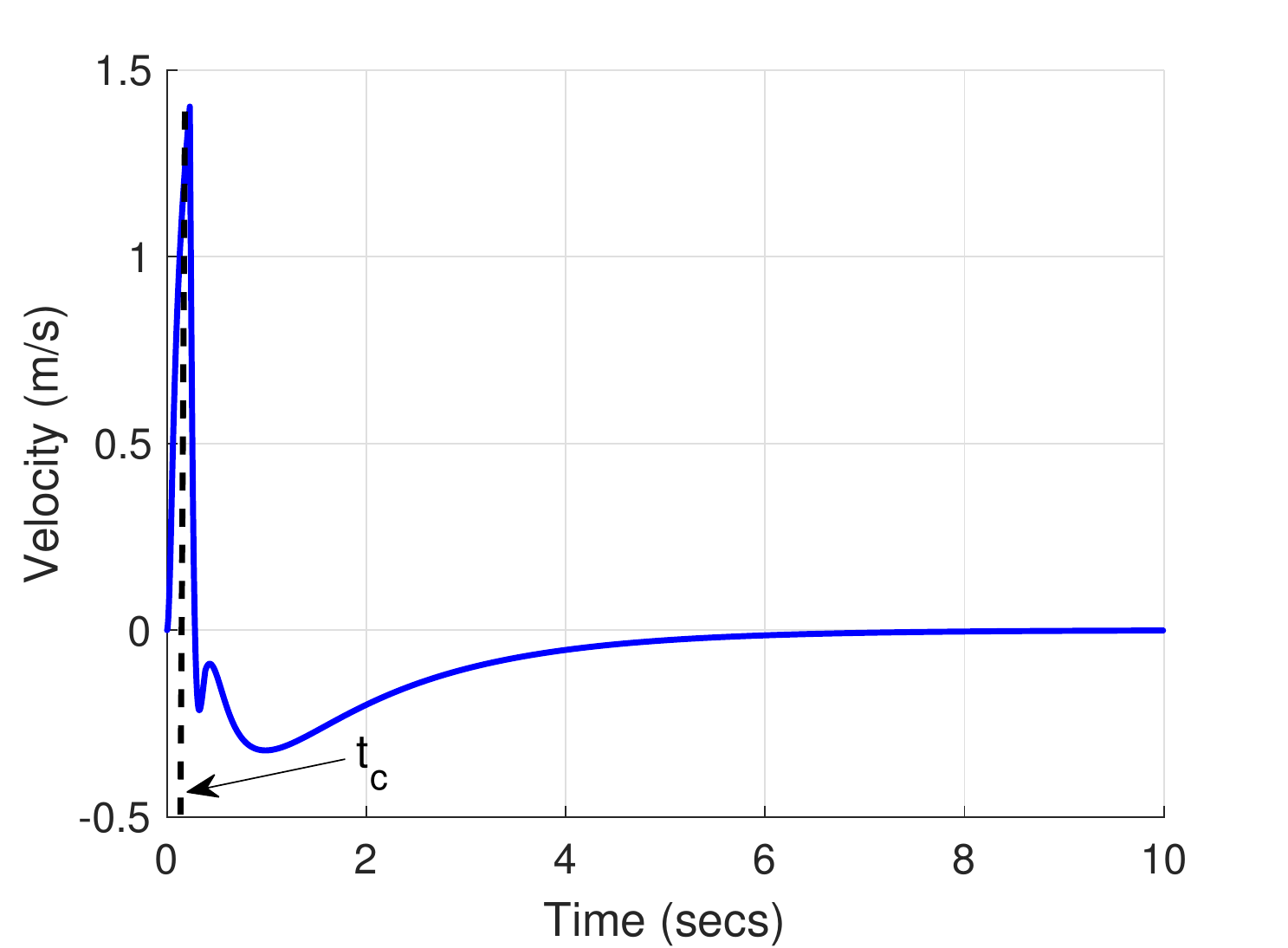}
		}
		\subfloat[{Enlarged view to highlight folding dynamics during collision} \label{fig:simxdot}]
		{\includegraphics[width = 0.23\textwidth]{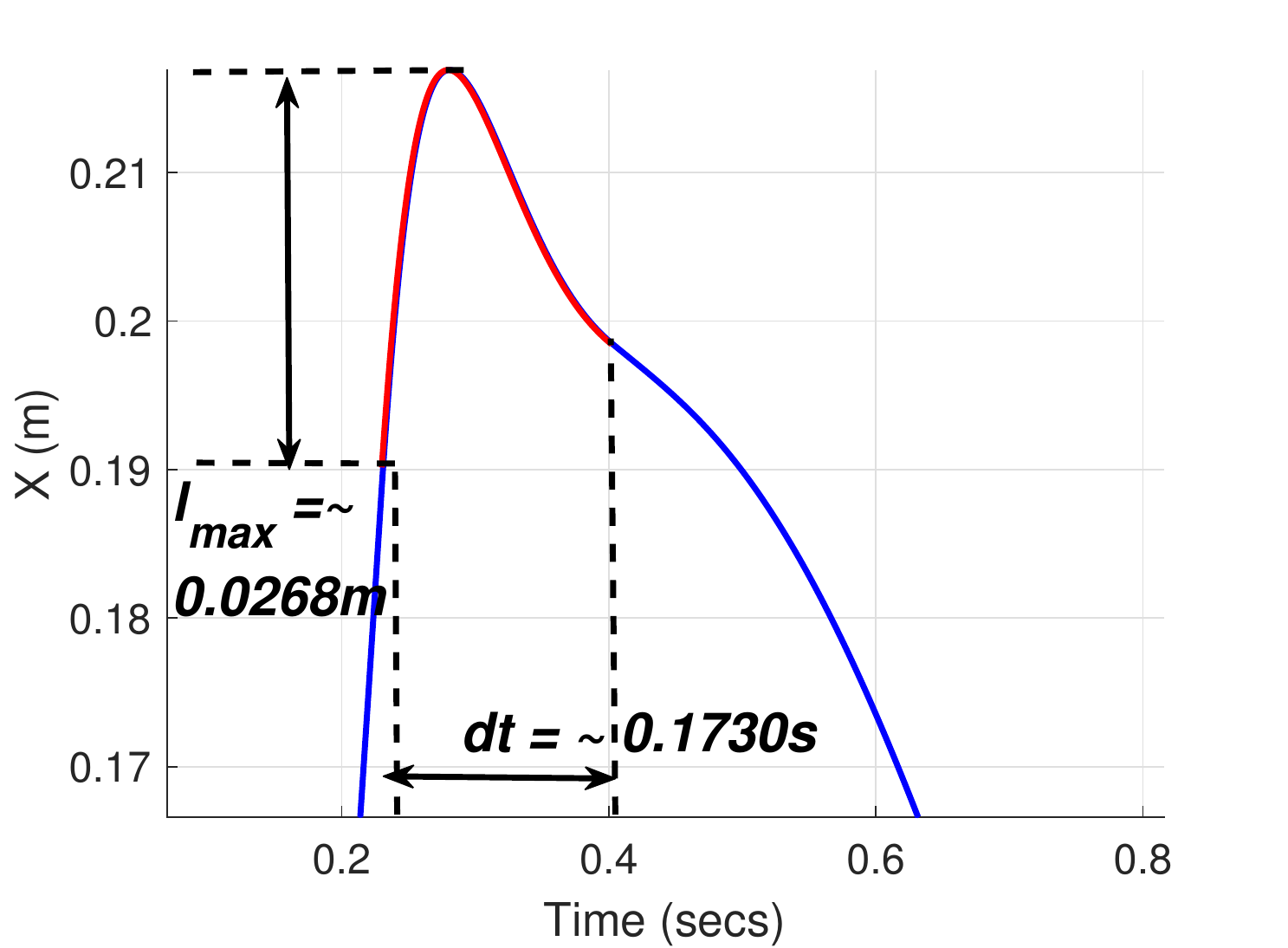}
		}
		\caption{Simulation results of the quadrotor trajectory in a sample collision scenario with a wall on the $i_1$ axis. (a) shows the parameter identification for  (\ref{eqn:spring_soln}). (b) and (c) show the position and velocity trajectory before and after collision, $x_d(t_c)$ is updated using (\ref{eqn:rebound}). The red segment in (b) and (d) depict the trajectory of the centroid of the quadrotor during the flight, highlighting the arm compression and release. $x=0$ denotes the origin of the inertial frame.}
		\label{fig:sim}
		\vspace{-0.2in}
	\end{figure}

	\section{Simulations and Experimental Validation}
	For the simulations, the parameters of the mass and the arm spring-damper system are set to $m = 1.112~ kg, ~J = diag[0.0034, ~0.0034, ~0.0053]^T, ~l_{arm} = 0.11 m, ~l_{max} = 0.03m, ~b_s = 30$ and $k_s = 500$ as experimentally obtained for the current foldable quadrotor. 
	The spring and damping coefficients are tuned to match the real system dynamics as shown in Fig. \ref{fig:coeffs}. A sample scenario is constructed where a wall is located in the $i_1$ axis $0.3m$ from the origin. The corresponding response is shown in Fig. \ref{fig:sim}. To ensure that the newly generated setpoint does not deviate significantly from $x(t_0)$. $\gamma_1$ is set to 0.5. In Fig. \ref{fig:simx}, the quadrotor is approaching a set point $x_d=[2,0,-4]^Tm$ without the knowledge of a wall ahead. It hits the wall at around $0.2s$, compresses against the wall and then bounces back. The red segment marked in Fig. \ref{fig:simx} shows the dynamics of the arm during collision which is zoomed in Fig. \ref{fig:simxdot}. When the quadrotor hits the wall, the arm moves inwards and consequently the centroid moves forward until it compresses completely against the wall and then comes backward as the arm comes back to its rest position. From Fig. \ref{fig:coeffs}, due to hysteresis, $l$ doesn't completely come back to zero and stays $\delta l$ away from zero. To account for this, we calculate $dt$ as the time difference between the instant when the quadrotor first hits the wall and time instant where $l \leq \delta l$ after $l$ reaches $l_{max}$. This corresponds to a slightly folding configuration while the quadrotor flies away as shown by the red marking in Fig. \ref{fig:simxdot}. And as described in Sec \ref{sec:icanalysis}, the quadrotor then switches to normal mode and flies off from the wall toward the new set-point. The corresponding velocity profile is shown in Fig. \ref{fig:simexv}. The pre-impact velocity is at approximately $1.4 m/s$ and the corresponding post-collision bounce off velocity is around $0.28 m/s$ which is significantly lower compared to a rigid quadrotor. 
	
\begin{figure}[t]
		\centering
		\subfloat[{Simulation: Position plot} ]
		{\includegraphics[width = 0.22\textwidth]{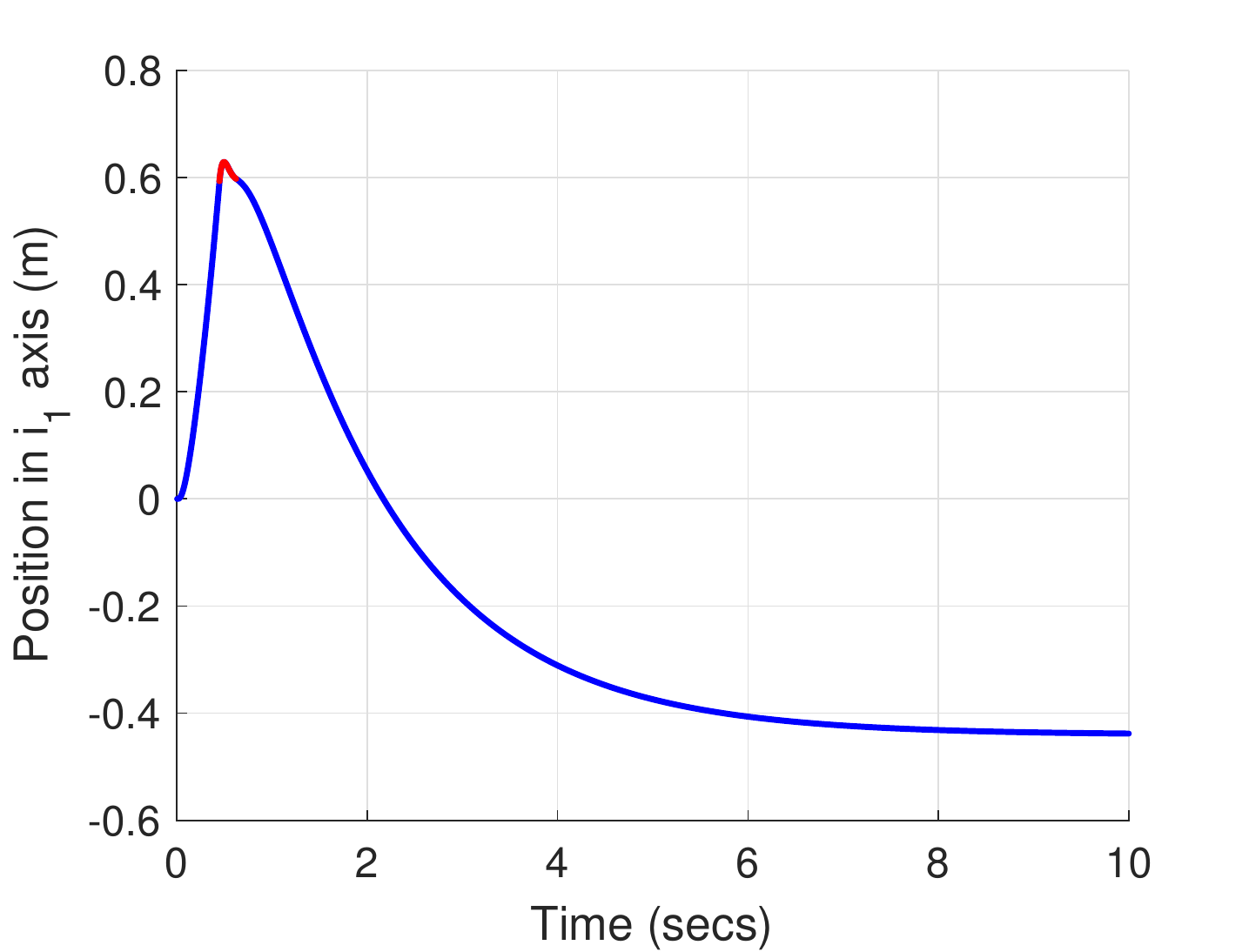}
		}
		\subfloat[{Simulation: Trajectory of $v_x$}]
		{\includegraphics[width = 0.23\textwidth]{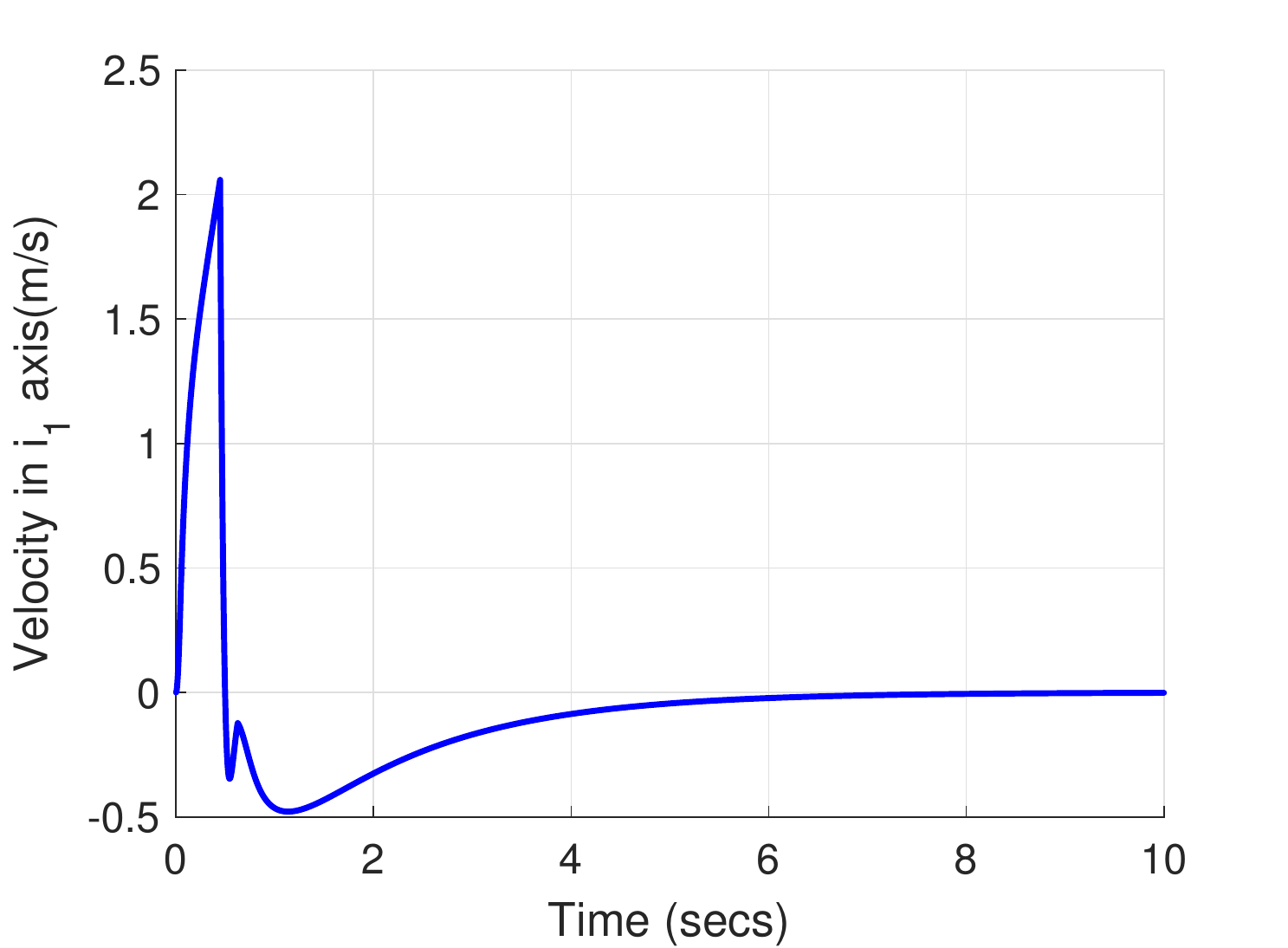}
		}\\
		\subfloat[{Experiments: Position plot}]
		{\includegraphics[width = 0.22\textwidth]{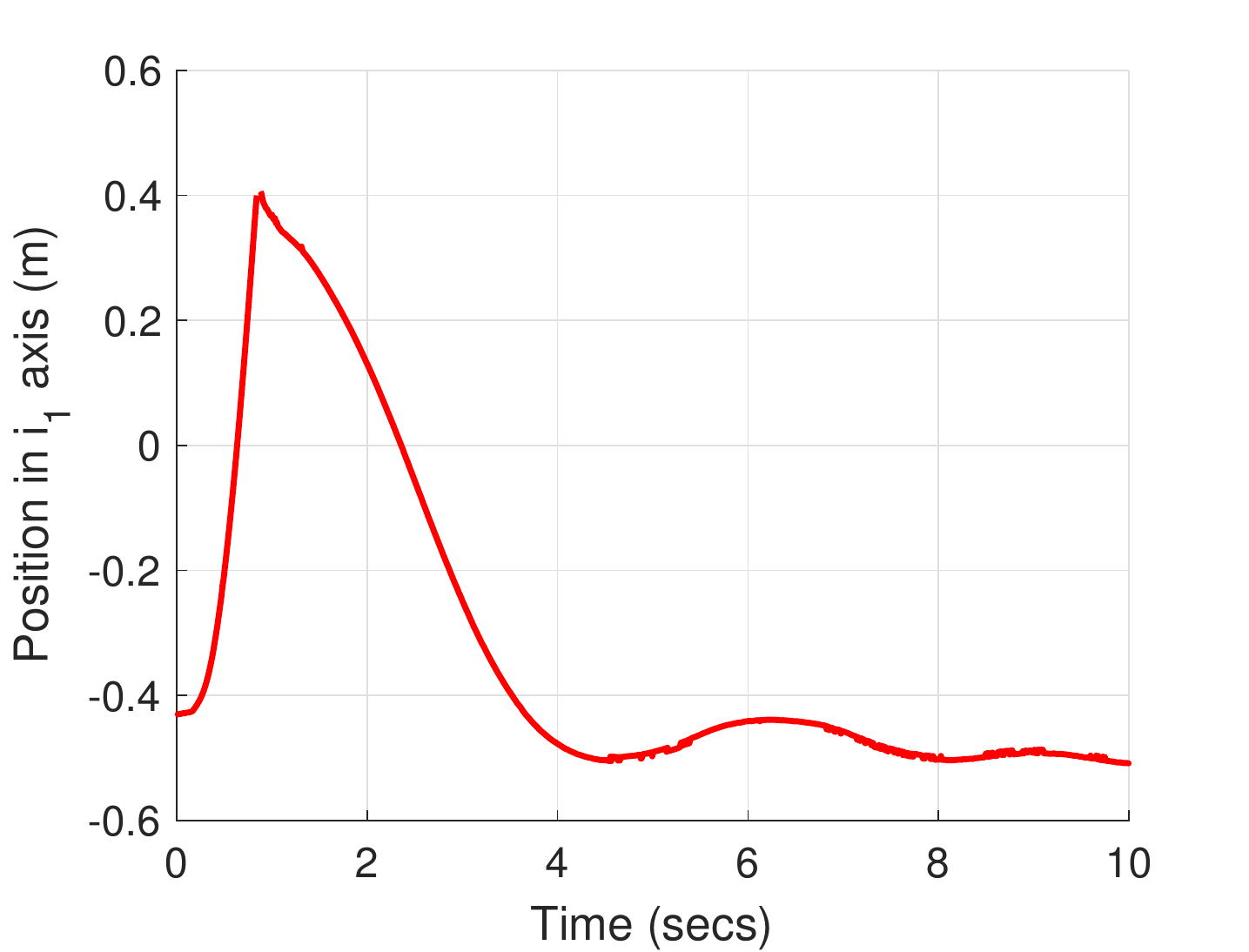}
		}
		\subfloat[{Experiments: Trajectory of $v_x$}]
		{\includegraphics[width = 0.23\textwidth]{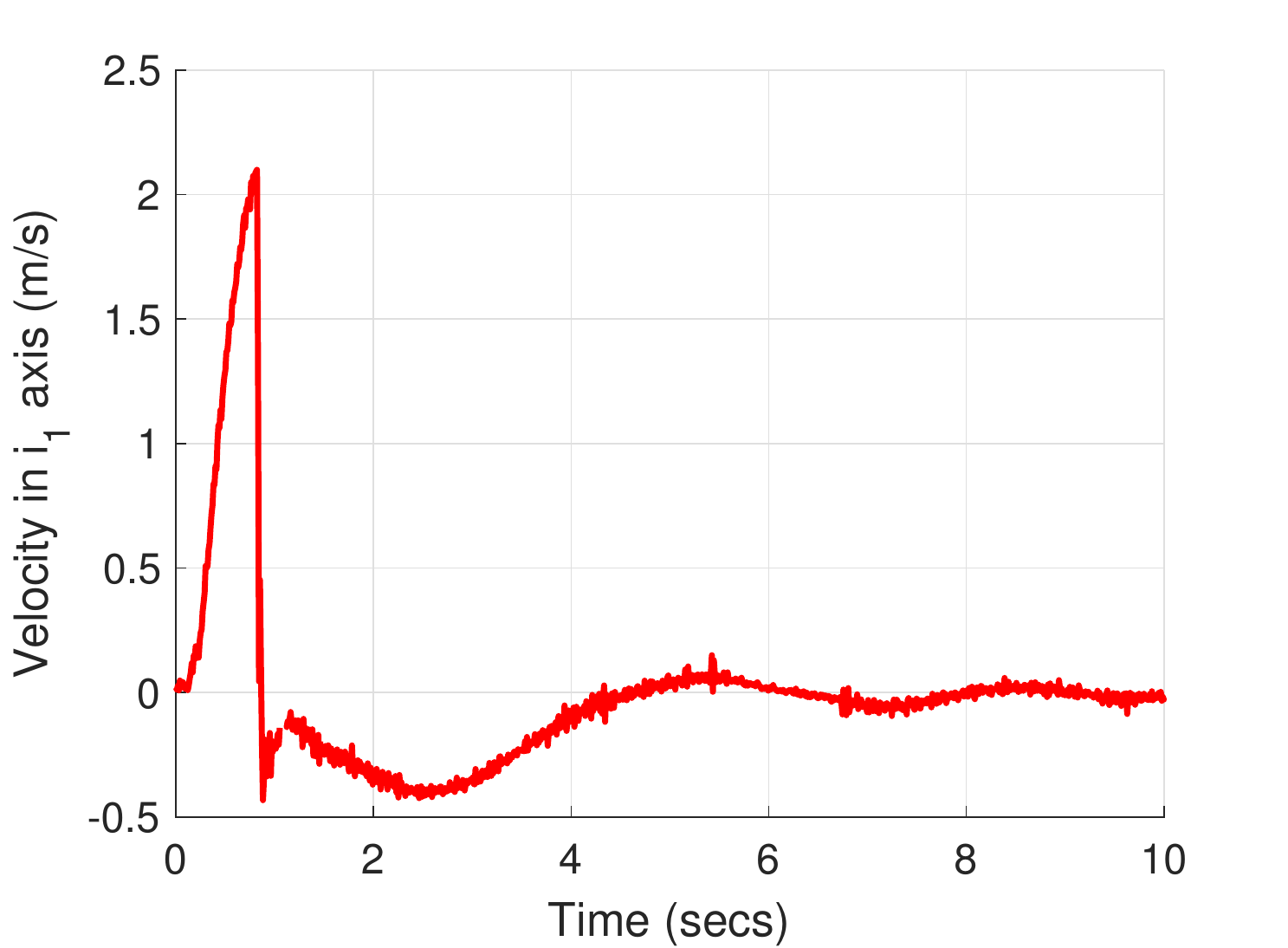}
		}
		\caption{Comparisons of simulation and experimental results for $v_c = 2.1 m/s$ and $\gamma = 0.5$, i.e $x_d(t_c)$ should be generated $1.05m$ away from $x(t_c)$. In (a) and (b), $x(t_c) = [0.6, ~0, ~-0.5]^T$ and $x_d(t_c) = [-0.45, ~0, ~-0.5]^T$. In (c) and (d), $x(t_c) = [0.4, ~0, ~-0.5]^T$ and from (\ref{eqn:rebound}), the $x_d(t_c) = [-0.65, ~0, ~-0.5]^T$. However in (d), $x_d(t_c) = [-0.55, ~0, ~-0.5]^T$ and this difference is attributed to the error in the measurement for $v_c$. All distances are in $m$.}
		\label{fig:sim_comparison}
		\vspace{-0.2in}
	\end{figure}

	
	The simulation results of the controller performance matched well with the experimental results for the foldable quadrotor, as shown in Fig. \ref{fig:sim_comparison} at a collision velocity of $2.1 m/s$, and $\gamma_1 = 0.5$ which confirm the hypothesis that the rebound velocity for the foldable quadrotor will be close to the one predicted as described in Sec. \ref{sec:icanalysis}. 

	\begin{figure*}[t]
		\centering
		\subfloat[\label{fig:posF1}{Position and velocity plot in $i_1$ axis for $v_c \approx 1~ m/s$ } ]
		{\includegraphics[width = 0.22\textwidth]{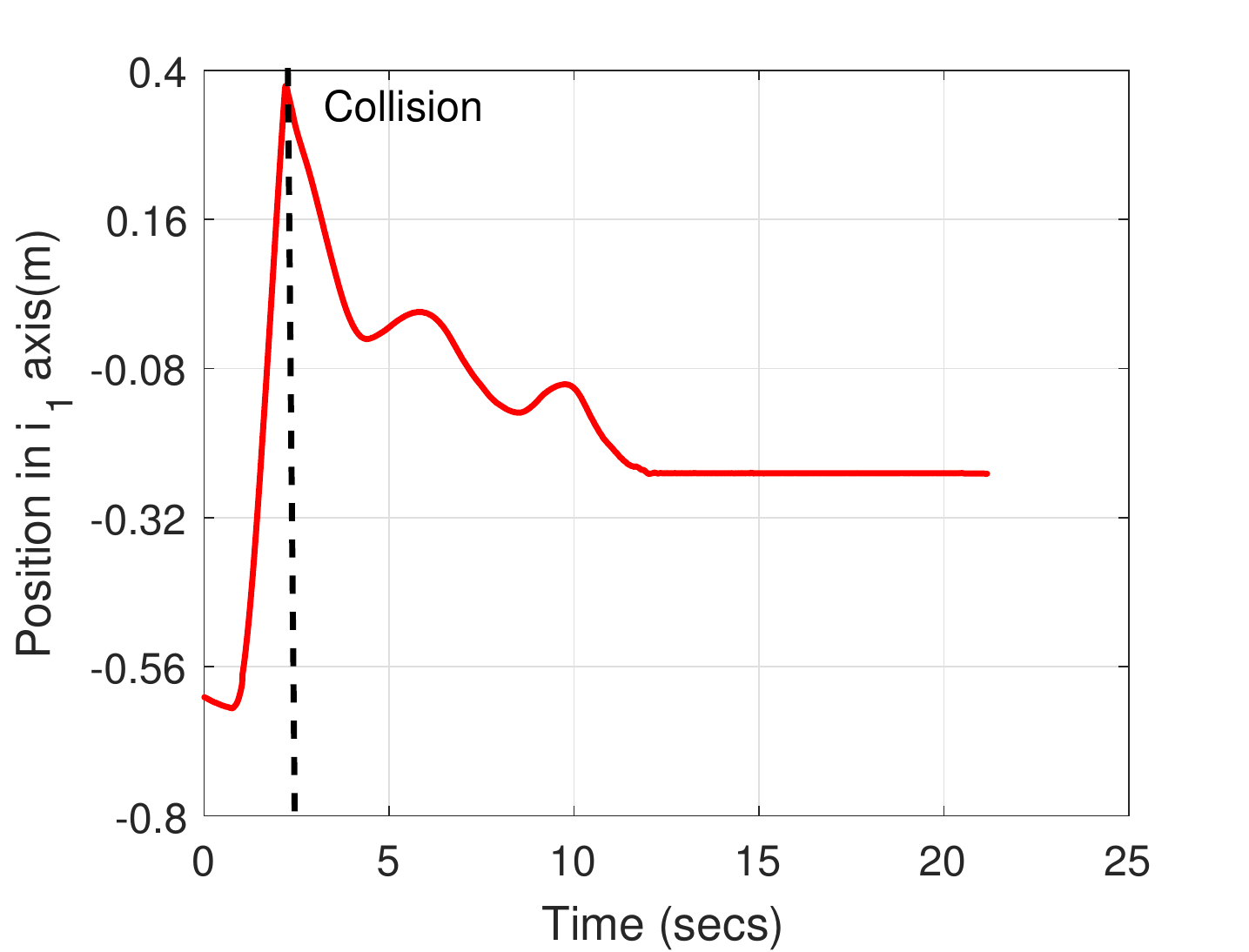}
			\includegraphics[width = 0.23\textwidth]{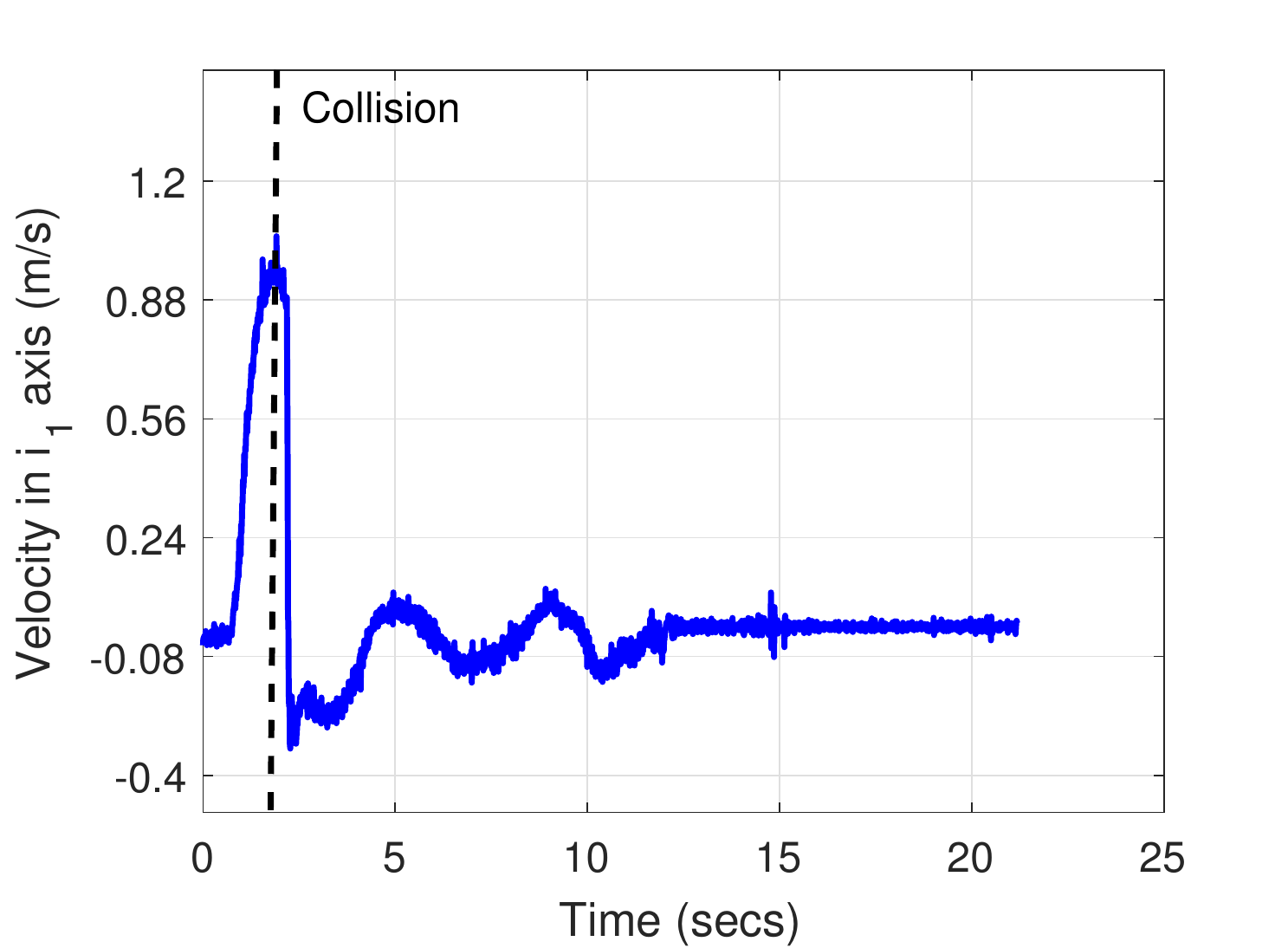}
		}
		\subfloat[\label{fig:f2}{Position and velocity plot in $i_1$ axis for $v_c \approx 1.5~ m/s$ } ]
		{\includegraphics[width = 0.22\textwidth]{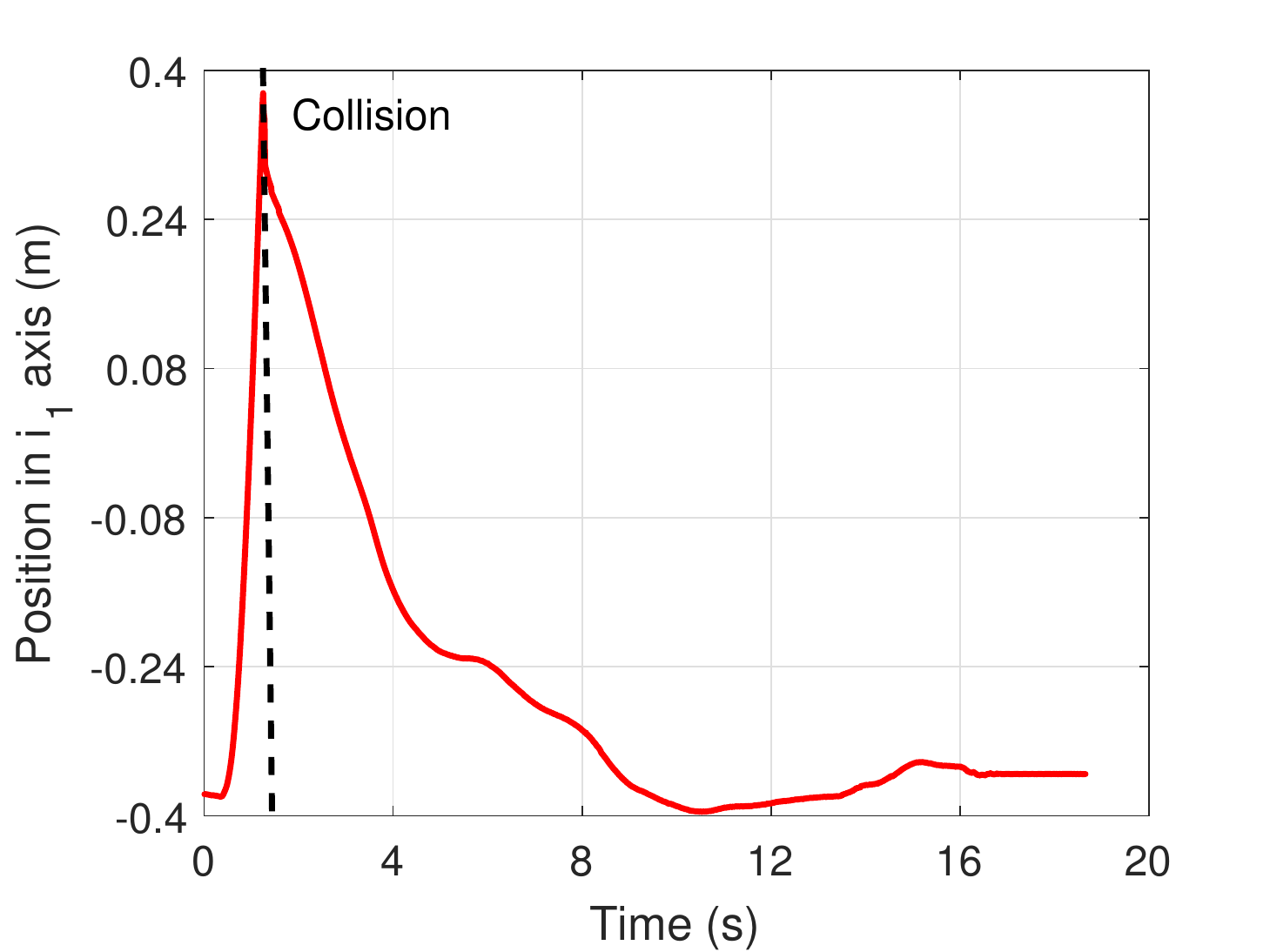}
			\includegraphics[width = 0.23\textwidth]{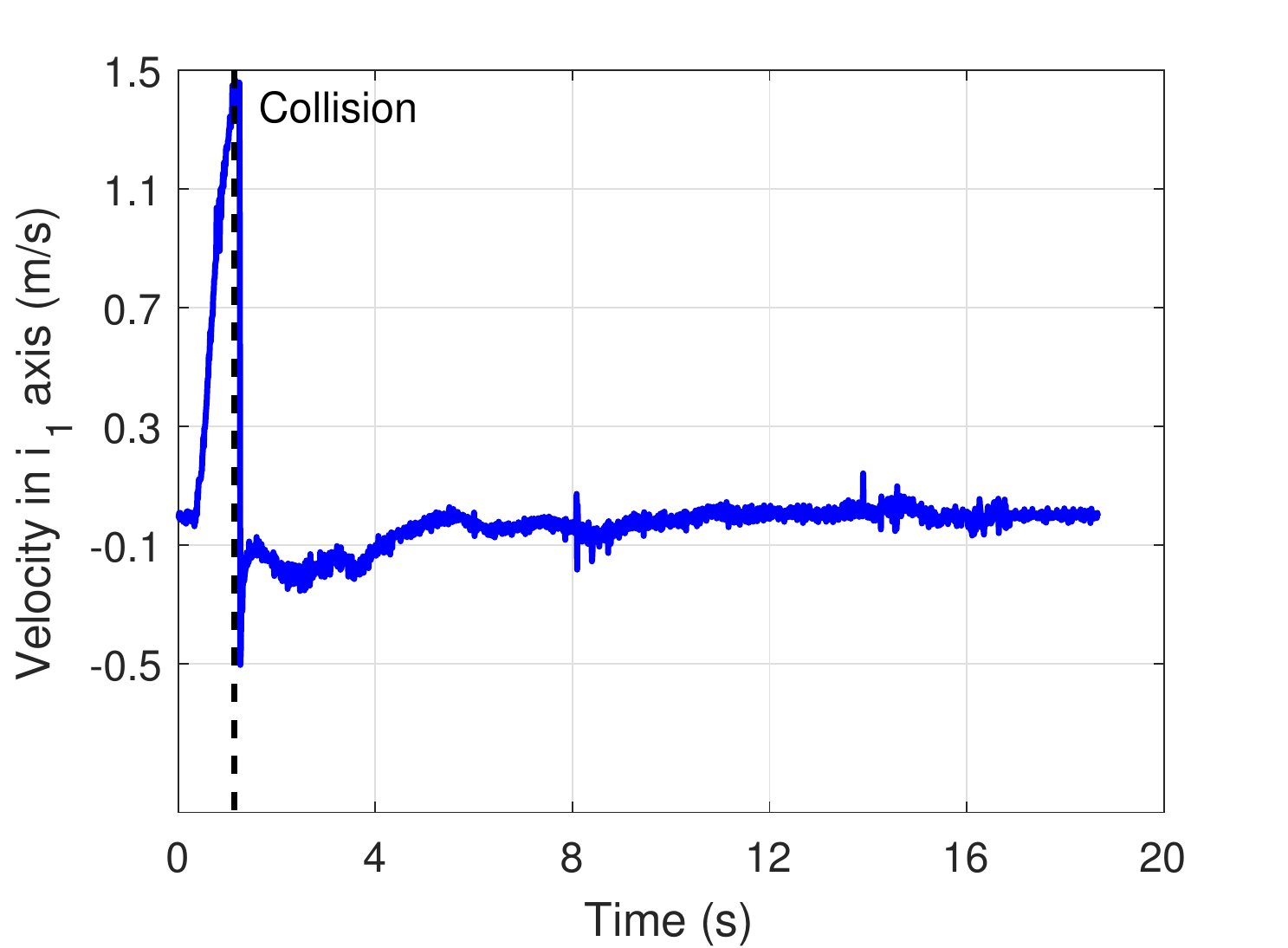}
		}\\
		\subfloat[\label{fig:f3}{Position and velocity plot in $i_1$ axis for $v_c \approx 2~ m/s$ } ]
		{\includegraphics[width = 0.22\textwidth]{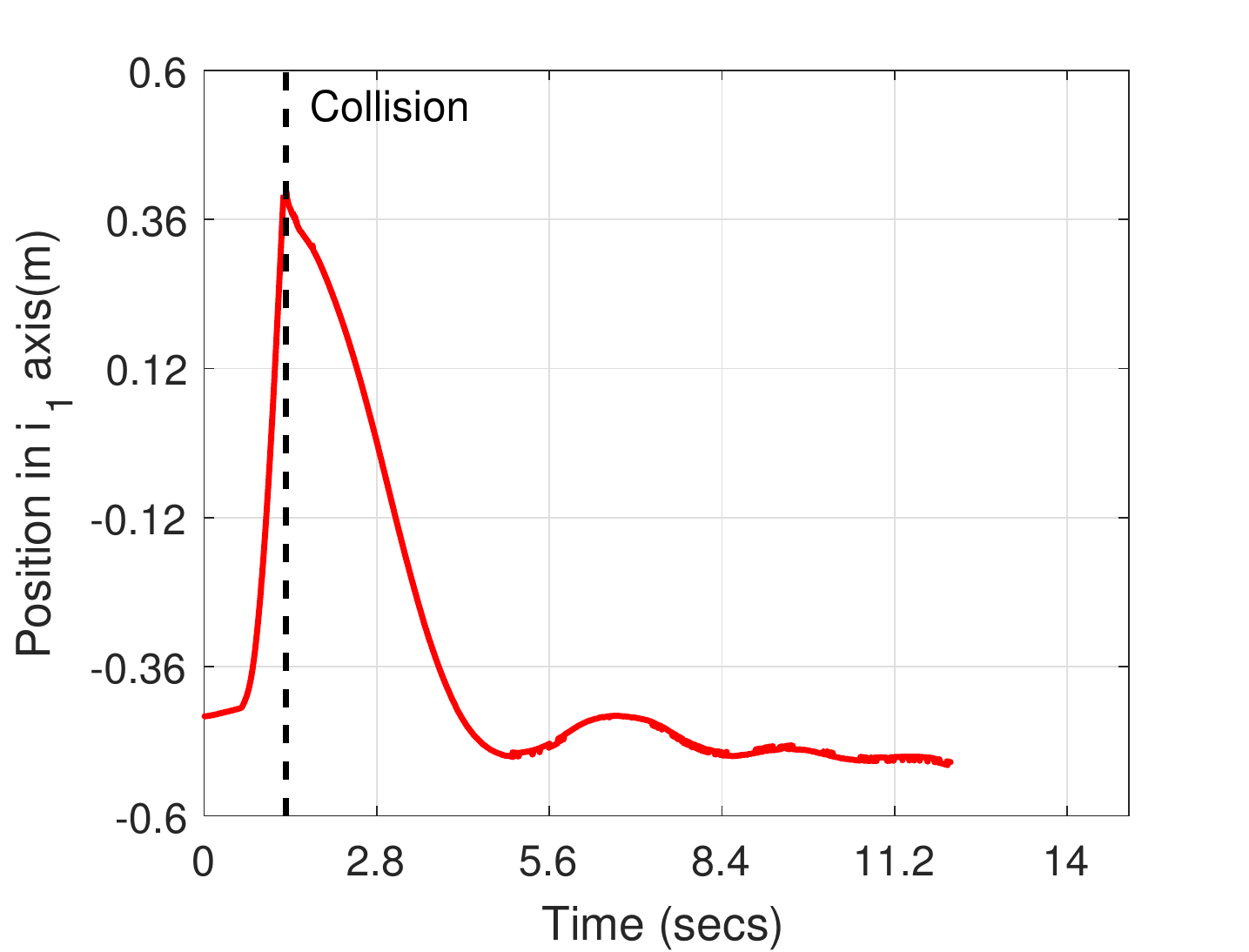}
			\includegraphics[width = 0.23\textwidth]{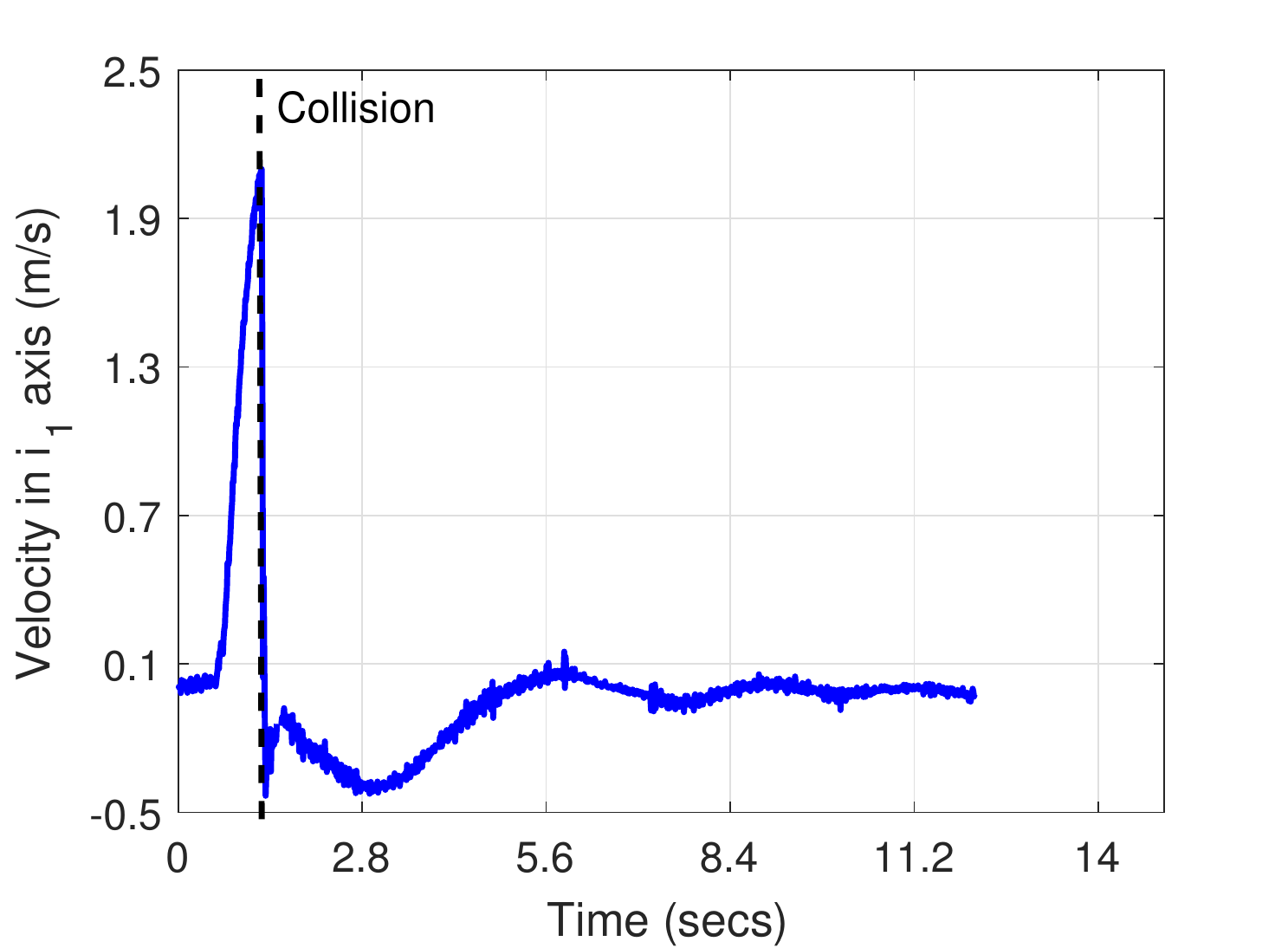}
		}
		\subfloat[\label{fig:f4}{ Position and velocity plot in $i_1$ axis for $v_c \approx 2.5~ m/s$}]
		{\includegraphics[width = 0.22\textwidth]{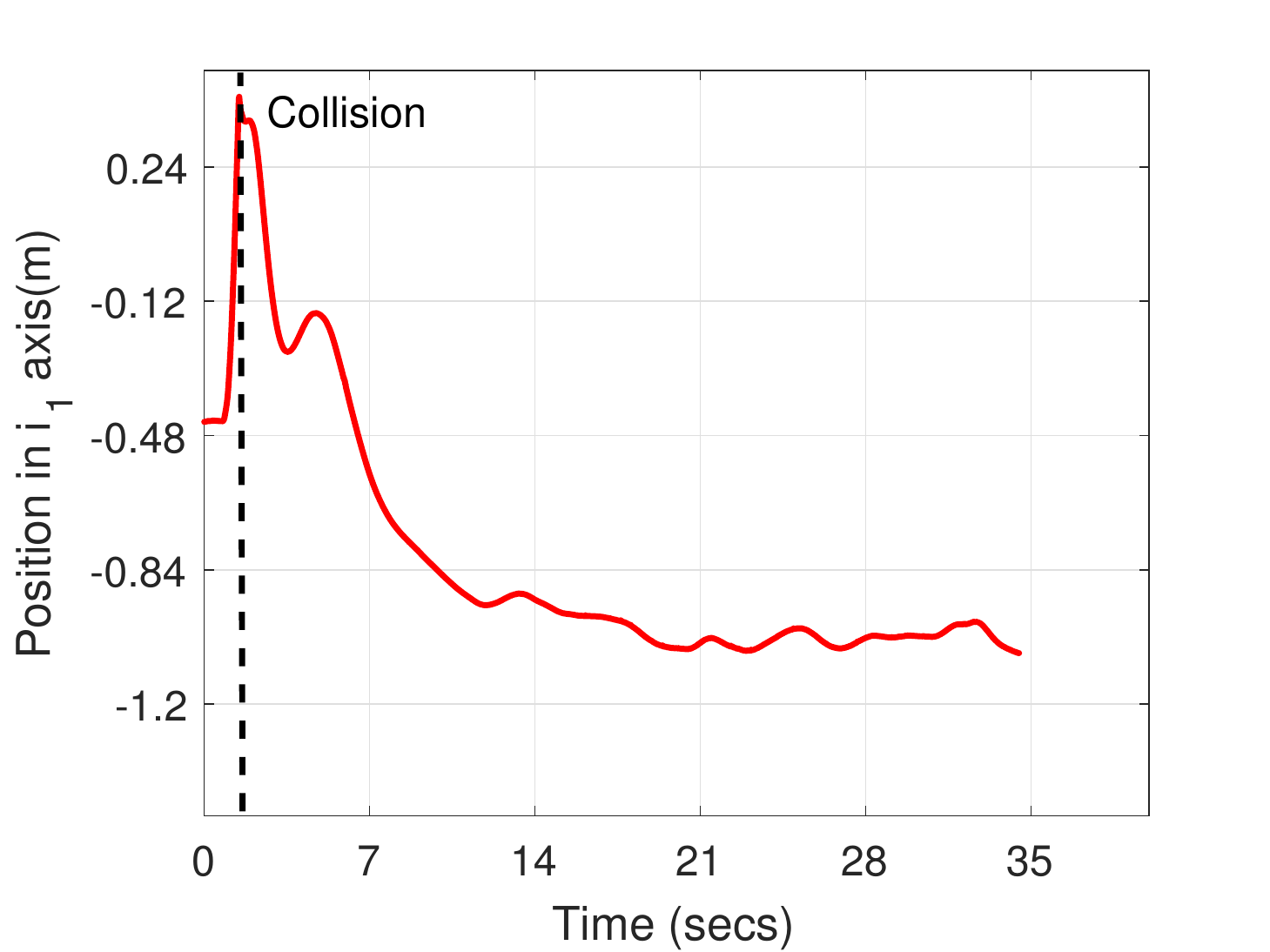}
			\includegraphics[width = 0.23\textwidth]{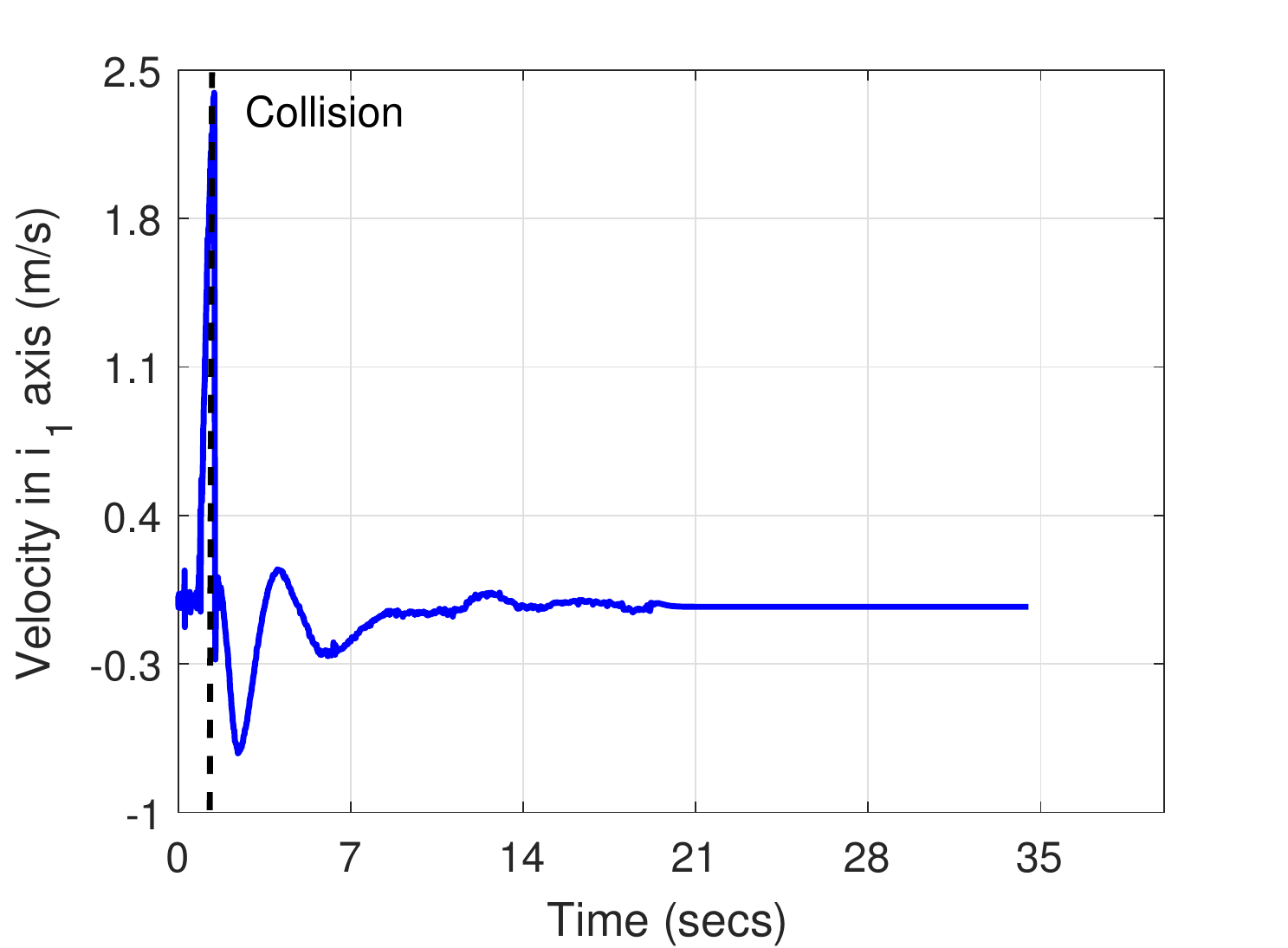}
		}
		\caption{The position and velocity trajectories of a foldable quadrotor for different scenarios. Collision velocity, $v_c$, ranges from 1-2.5 $m/s$. It is evident that after collision, the proposed quadrotor demonstrates accurate tracking of the new recovery setpoint with no overshoot for every case scenario.}
		\label{fig:exp_results1}
		\vspace{-0.2in}
	\end{figure*}
	\begin{figure*}
		\centering
		\subfloat[\label{fig:posR1}{Position and velocity plot in $i_1$ axis for $v_c \approx 1~ m/s$} ]
		{\includegraphics[width = 0.22\textwidth]{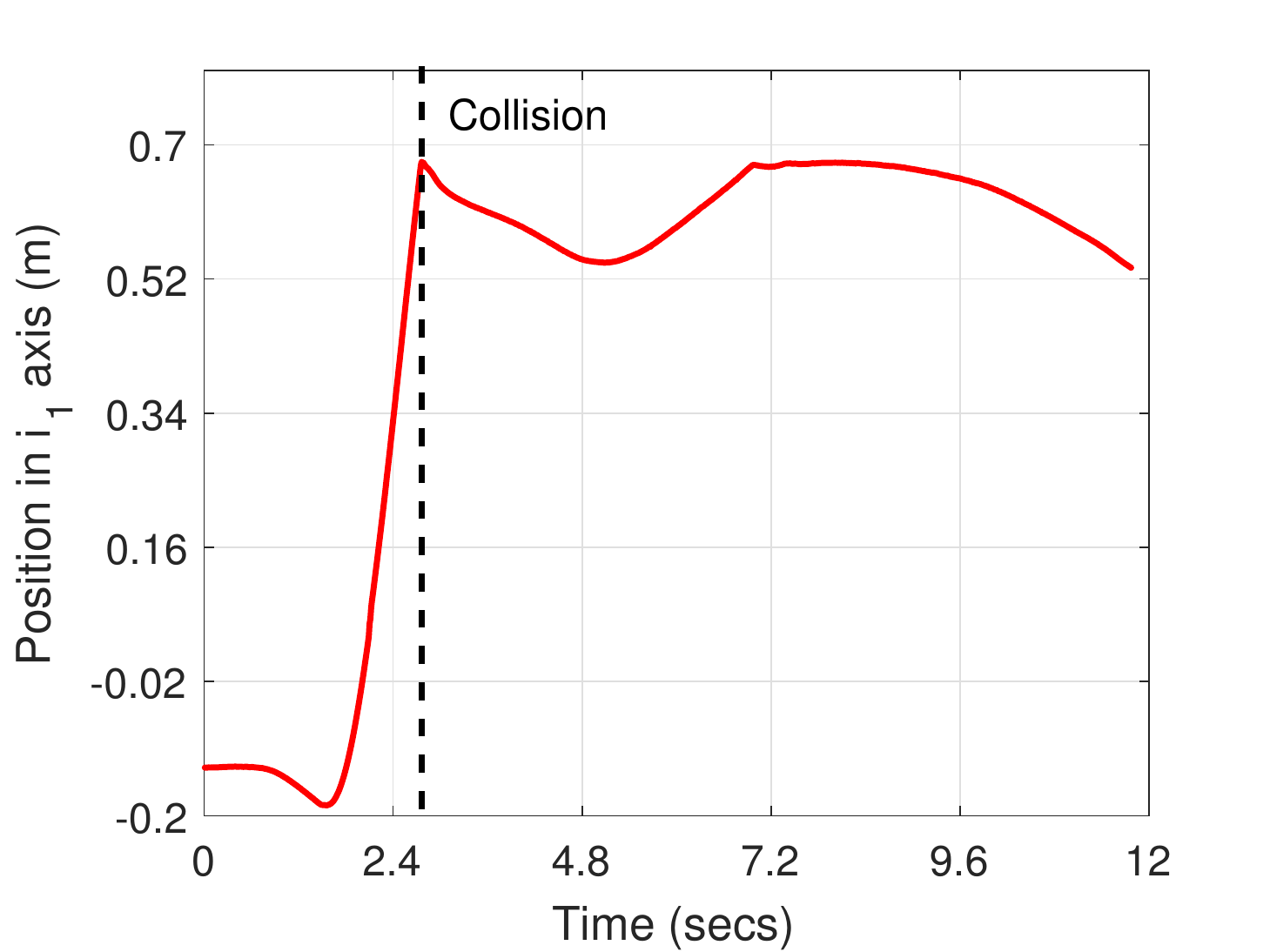}
			\includegraphics[width = 0.23\textwidth]{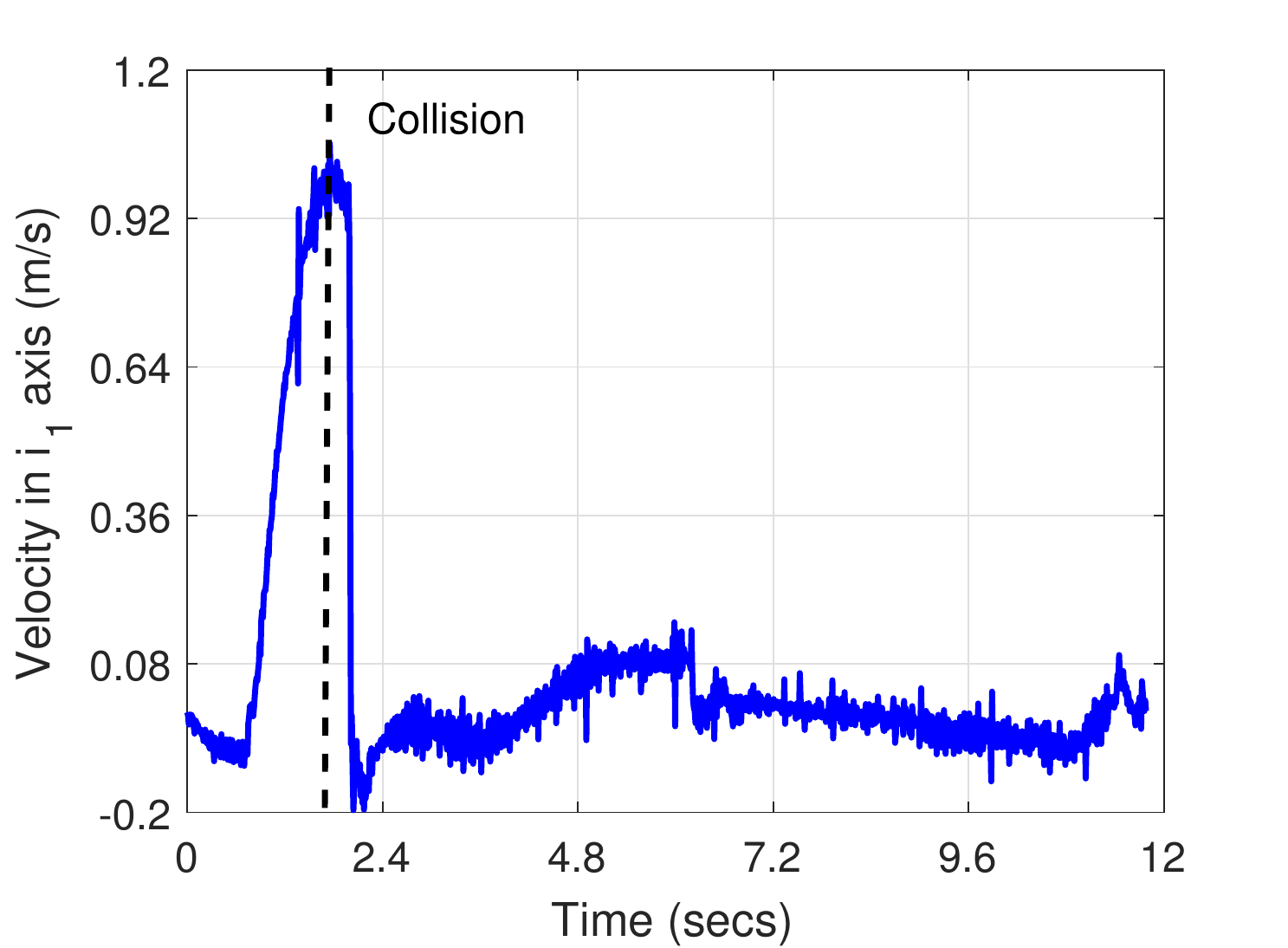}
		}
		\subfloat[\label{fig:vel15}{ Position and velocity plot in $i_1$ axis for $v_c \approx 1.5~ m/s$ } ]
		{\includegraphics[width = 0.22\textwidth]{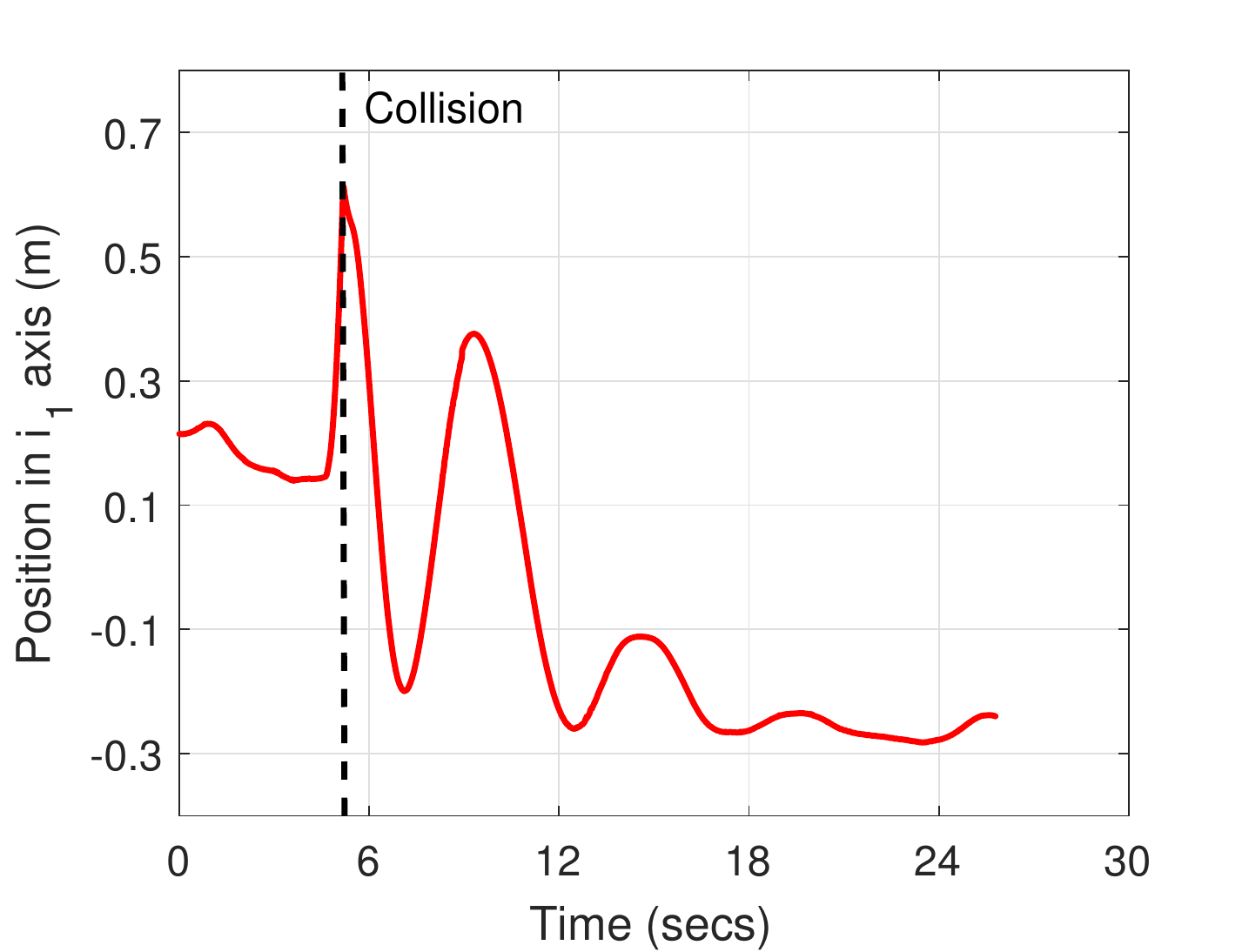}
			\includegraphics[width = 0.23\textwidth]{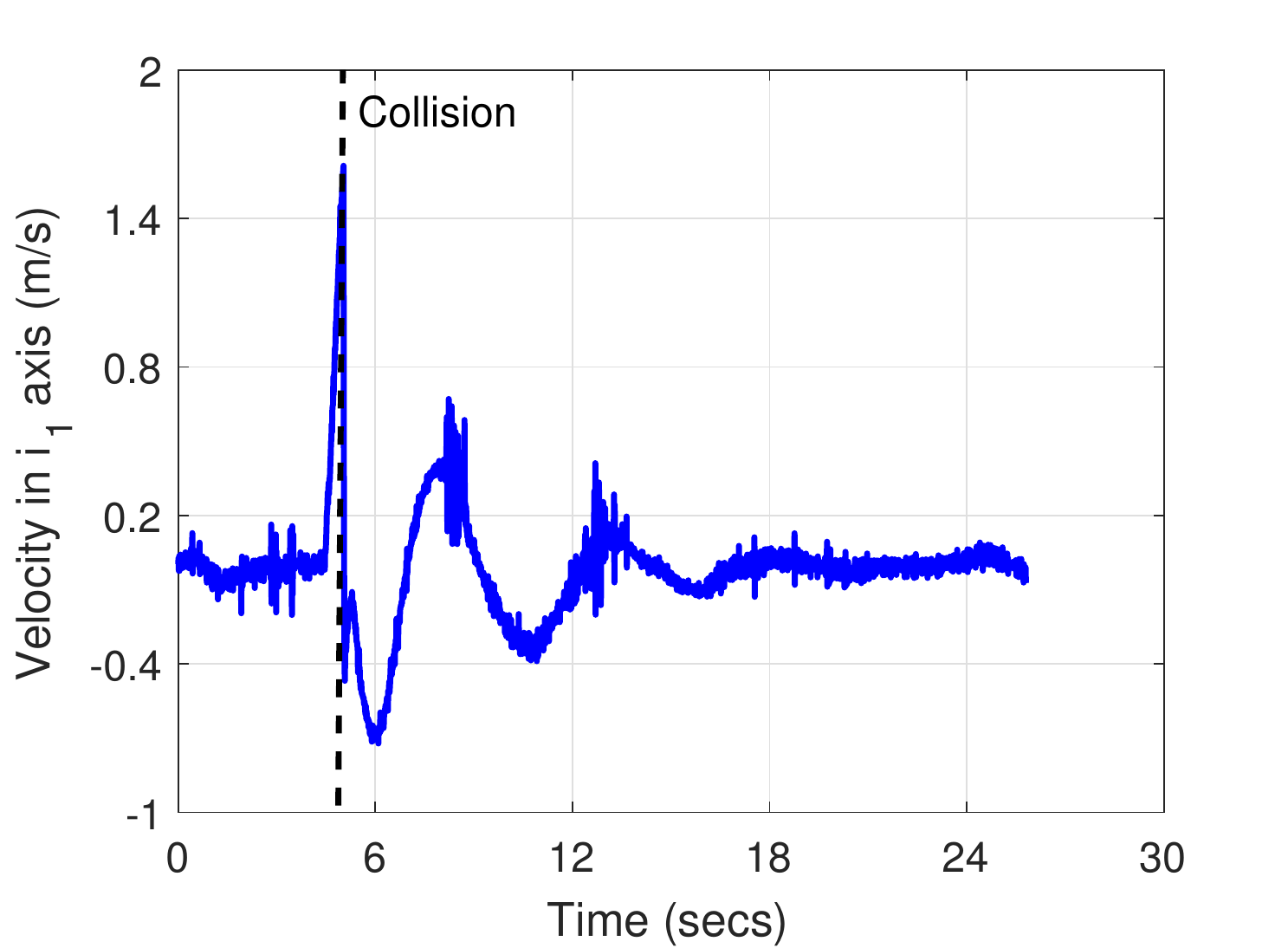}
		}\\
		\subfloat[\label{fig:velR3}{Position and velocity plot in $i_1$ axis for $v_c \approx 2~ m/s$} ]
		{\includegraphics[width = 0.22\textwidth]{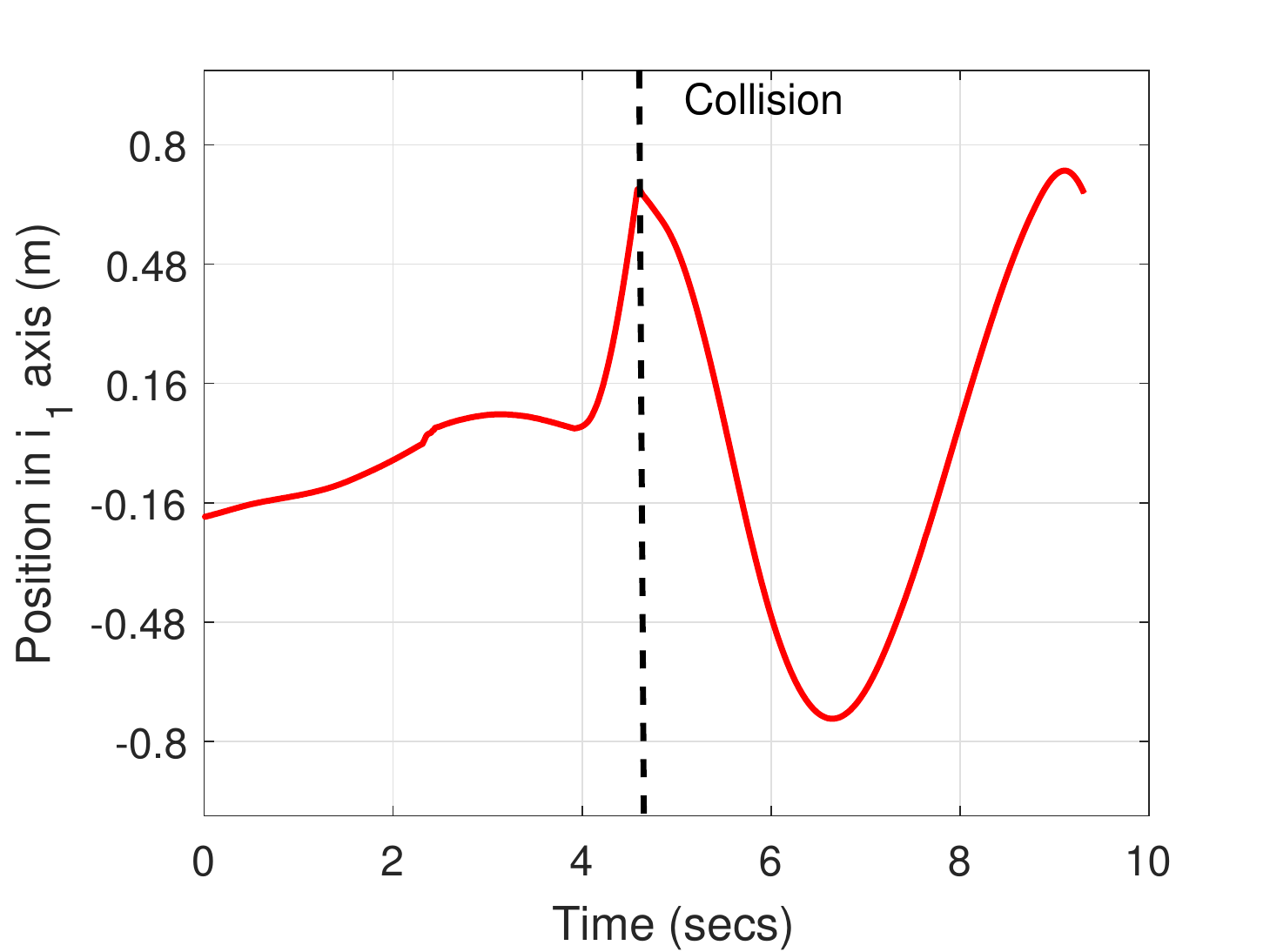}
			\includegraphics[width = 0.23\textwidth]{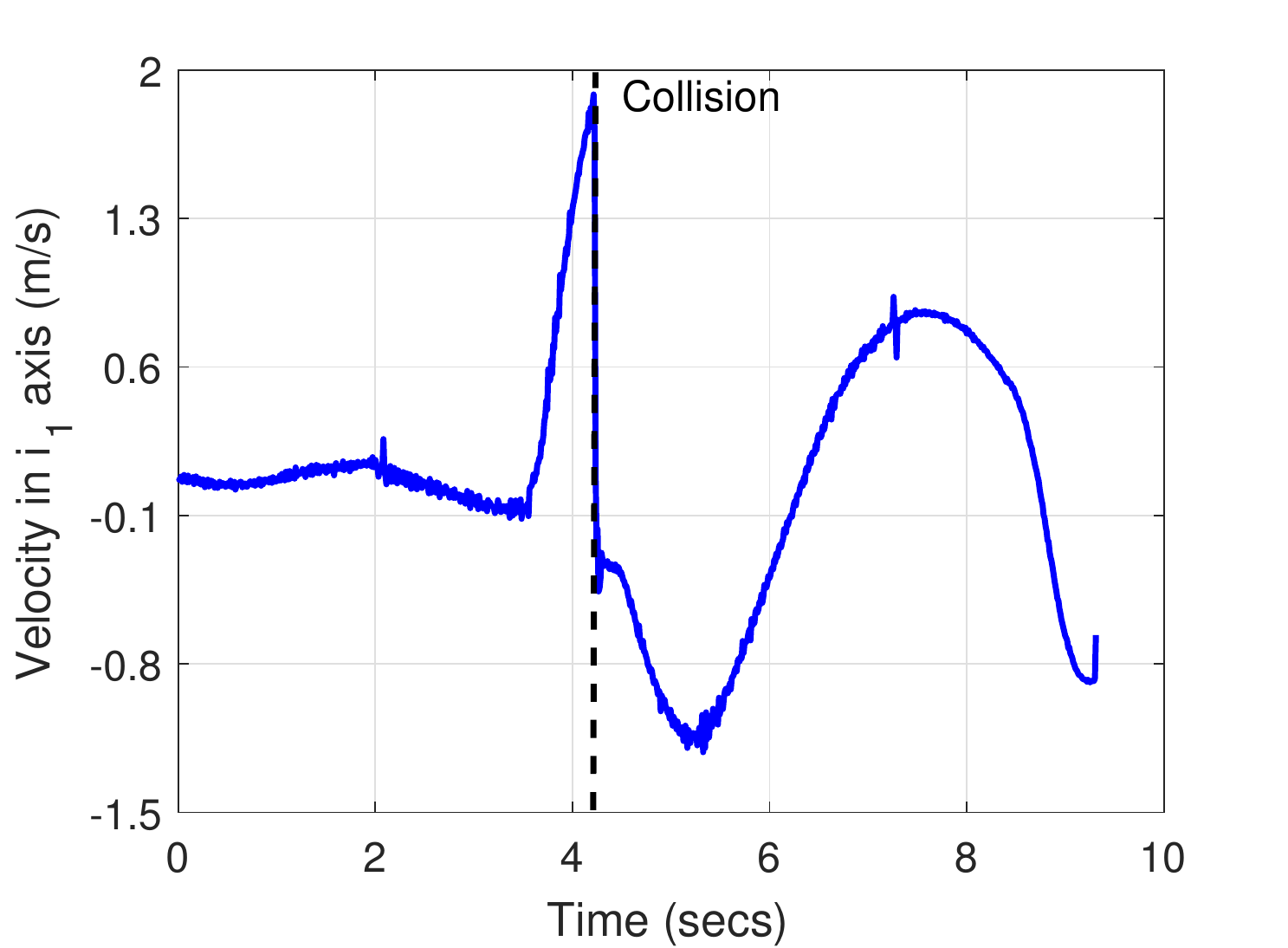}
		}
		\subfloat[{Position and velocity plot in $i_1$ axis for $v_c \approx 2.5~ m/s$}]
		{\includegraphics[width = 0.22\textwidth]{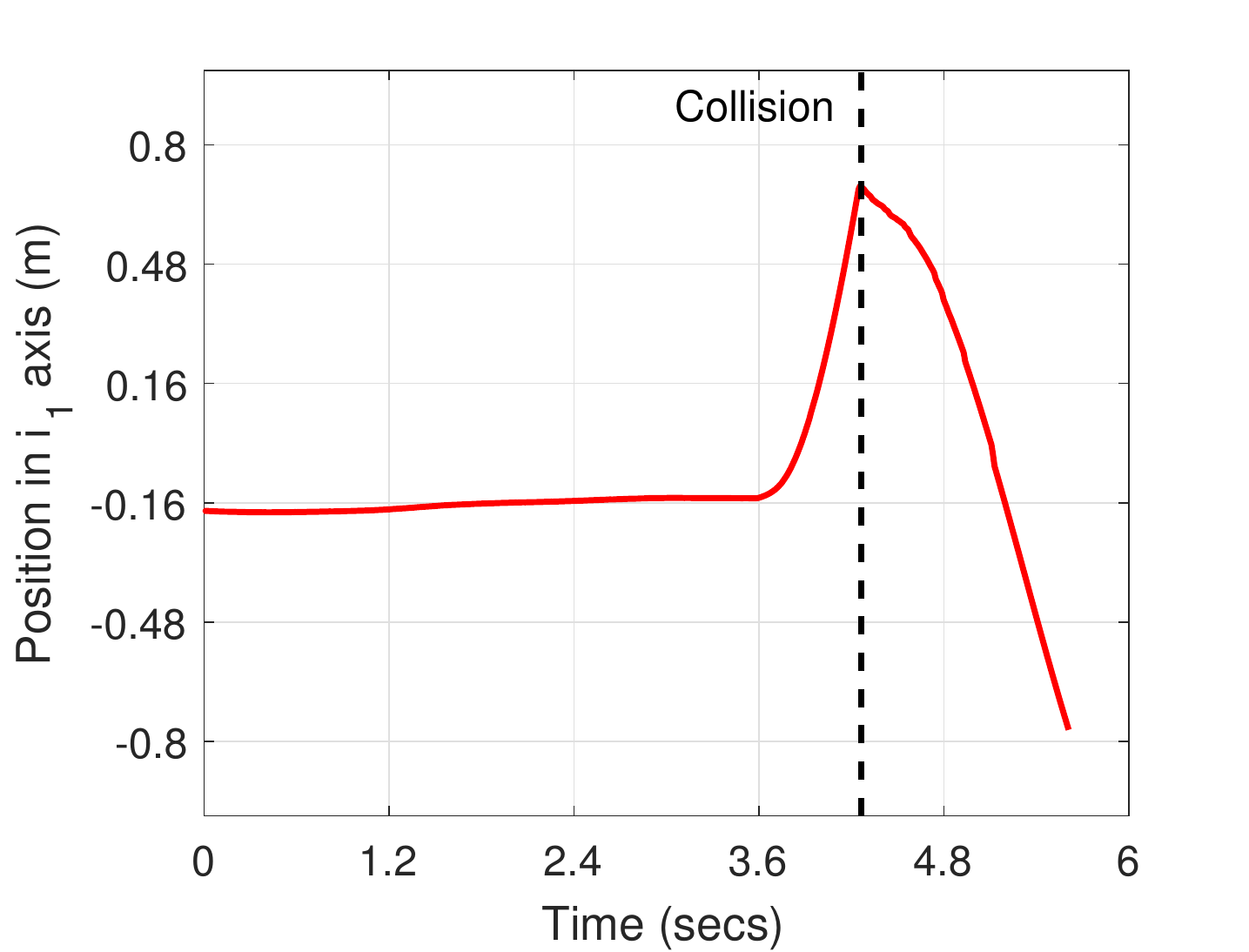}
			\includegraphics[width = 0.23\textwidth]{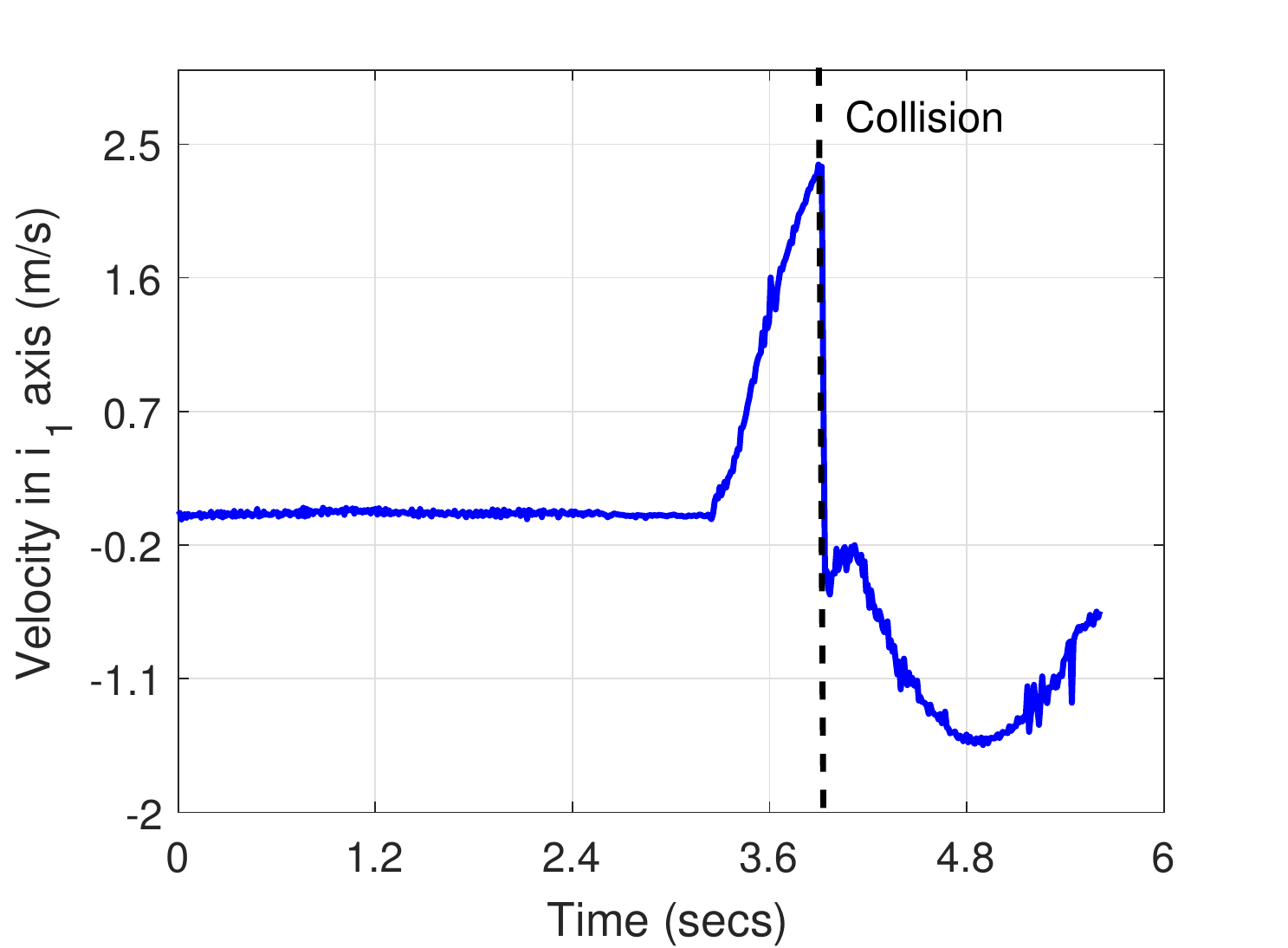}
		}
		\caption{The position and velocity trajectories of a rigid quadrotor for similar collision scenarios. The rigid quadrotor demonstrates unpredictable response for every scenario owing to various factors such as sample loss from the IMUs due to low impact times and oscillatory response due to high rebound velocities.}
		\label{fig:exp_results4}
		\vspace{-0.2in}
	\end{figure*}
	\section{Real-Time Flight Tests}
	This section describes the experimental setup and the flight tests performed for demonstrating the efficacy of the proposed mechanical design and control structure.
	\subsection{Hardware Setup}
	Multiple collision flight tests were conducted to compare the collision recovery performance between the proposed foldable quadrotor and a rigid quadrotor with a similar size, weight and material. The PIXHAWK was used as the flight control unit and the PX4FLOW was utilized for obtaining real-time velocity information. A single point-detection distance sensor (LeddarOne, LeddarTech, Quebec City, Canada) was utilized for altitude control. An extended Kalman filter was used to estimate the orientation and position of the quadrotor. A motion capture system (Vicon Motion Systems Ltd, Oxford, UK) was used to record the ground truth position and velocity data. 
	
	\subsection{Experimental Results}
	Real-time implementation of the recovery control strategy is done on the flight control unit, where the orientation control loop runs at about 150 Hz and position control loop runs at about 100 Hz. It is to be noted that the high impact times of the foldable quadrotor allow for the use of onboard IMUs to detect collision consistently and generate the new setpoint unlike in the rigid case, where the collision detection can fail if the sample time is higher than the impact time. The value of the estimated velocity at the time of collision is used in (\ref{eqn:rebound}) to generate the new setpoint. Although, the foldable quadrotor is designed to endure impact velocities of up to $5m/s$, to emulate realistic mission related velocities, the flight tests are carried out at velocities ranging from $1$ to $2.5m/s$.
	The new desired setpoint for both the systems is generated using $\gamma = 0.5$ in (\ref{eqn:rebound}). Fig. \ref{fig:posF1} shows the position and velocity trajectory respectively for the entire flight of the foldable quadrotor for $v_c = 1 ~m/s$, while Fig. \ref{fig:posR1} shows the comparison plot of the rigid system for the same $v_c$. It is noted from the Fig. \ref{fig:posF1} that the foldable quadrotor's low rebound velocity and longer impact time ($\approx0.025s$), enables successful tracking of the post collision recovery setpoint with no overshoot. The inward movement of the arm during spring compression also contributes to a favorable response by constraining majority of the impact force in the direction of the compression. However, in the rigid quadrotor's position plot of Fig. \ref{fig:posR1}, we see that high rebound velocity and very low impact times ($\approx0.001s$), adversely affect the performance of the recovery control, leading to an unpredictable, oscillating response. A video demonstration validating the superior recovery performance of the proposed platform can be found at \url{https://youtu.be/obepryCvdZ4} while the rigid one has an unpredictable response. Similar results obtained for other velocities verify the consistent and reliable performance of the proposed design as shown in Figs. \ref{fig:f2}, \ref{fig:f3} and \ref{fig:f4}. Additionally, it is evident that with the increase in impact velocities, there is an increase in the amplitude of the oscillations for the rigid quadrotor. Therefore, the rigid quadrotor may overshoot the new setpoint for higher impact velocities and rerun into the wall, as seen in Fig. \ref{fig:velR3}.  
	
	
	\section{Conclusion and Future Work}
	In this paper, a collision recovery control with a foldable quadrotor was presented. An FEA analysis was performed to analyze the improvement of the factor of safety as compared to a conventional quadrotor. Additionally, the dynamic model for the system during the collision was derived and analysed to develop a recovery controller. Experimental tests demonstrated that the foldable quadrotor could sustain collisions at different impact velocities due to the coupled advantage of a compliant mechanical design and the recovery control algorithm. Comparison experiments with a rigid quadrotor showed promising results for the new foldable design. 
	\par
	Future work includes improving the mechanical flexibility of the current frame to sustain collisions from any direction. Collision recovery techniques including detailed collision characterization will be explored to account for multiple contacts and improve altitude tracking.

	
	\printbibliography

@inproceedings{BP13,
  title={Contact-based navigation for an autonomous flying robot},
  author={Broid \textit{et al}, A.},
  booktitle={Int. Conf. Intelligent Robot Systems},
  pages={201-213},
  organization={IEEE},
  year ={2013 }
}

@inproceedings{P+20,
  title={Design and Control of SQUEEZE: A Spring-augmented QUadrotor for intEractions with the Environment to squeeZE-and-fly},
  author={Patnaik \textit{et al}, K },
  booktitle={Int. Conf. Intelligent Robots and Systems},
  pages = {1364-1370},
  organization={IEEE},
  year ={2020 }
}

@inproceedings{BS12,
  title={A hybrid pose/wrench control framework for quadrotor helicopters},
  author={Bellens, Steven and De Schutter, Joris and Bruyninckx, Herman},
  booktitle={IEEE Int. Conf. Robotics and Automation},
  pages={2269--2274},
  year={2012},
  organization={IEEE}
}

@article{RC15,
  title={Passivity-based control of VToL UAVs with a momentum-based estimator of external wrench and unmodeled dynamics},
  author={Ruggiero \textit{et al}, F. },
  journal={Robotics and Autonomous Systems},
  volume={72},
  pages={139--151},
  year={2015},
  publisher={Elsevier}
}

@inproceedings{KB13,
  title={Euler spring collision protection for flying robots},
  author={Klaptocz, Adam and Briod, Adrien and Daler, Ludovic and Zufferey, Jean-Christophe and Floreano, Dario},
  booktitle={Int. Conf. Intelligent Robots and Systems},
  pages={1886--1892},
  organization={IEEE},
  year = {2013 }
}

@inproceedings{TH14,
  title={A unified framework for external wrench estimation, interaction control and collision reflexes for flying robots},
  author={Tomi{\'c}, Teodor and Haddadin, Sami},
  booktitle={Int. Conf. Intelligent Robots and Systems},
  pages={4197--4204},
  organization={IEEE},
  year = {2014 }
}

@inproceedings{LM19,
  title={Design, Planning, and Control of an Origami-inspired Foldable Quadrotor},
  author={Yang \textit{et al}, Dangli},
  booktitle={American Control Conf.},
  pages={2551--2556},
  organization={IEEE},
  year = {2019 }
}

@article{OS12,
  title={Unmanned aerial vehicle (UAV) for monitoring soil erosion in Morocco},
  author={d'Oleire-Oltmanns \textit{et al}},
  journal={Remote Sensing},
  volume={4},
  number={11},
  pages={3390--3416},
  year={2012},
  publisher={Molecular Diversity Preservation Int.}
}

@inproceedings{NL13,
  title={Hybrid force/motion control and internal dynamics of quadrotors for tool operation},
  author={Nguyen, Hai-Nguyen and Lee, Dongjun},
  booktitle={Int. Conf. Intelligent Robots and Systems},
  pages={3458--3464},
  year  = {2013},
  organization={IEEE}
}

@article{C01,
  title={Impact dynamics and damage in composite structures},
  author={Christoforou, A. P.},
  journal={Composite structures},
  volume={52},
  number={2},
  pages={181--188},
  year={2001},
  publisher={Elsevier}
}

@inproceedings{ML18,
  title={Design and Control of a Hexacopter With Soft Grasper for Autonomous Object Detection and Grasping},
  author={Mishra \textit{et al}, S.},
  booktitle={ASME Dynamic Systems and Control Conf.},
  organization = {ASME},
  pages = {V003T36A003},
  year={2018}
}

@article{DZ95,
  title={Impact damage prediction in carbon composite structures},
  author={Davies, G. and Zhang, X.},
  journal={Int. Journal of Impact Engineering},
  volume={16},
  number={1},
  pages={149--170},
  year={1995},
  publisher={Elsevier}
}

@article{BC11,
  title={The navigation and control technology inside the ar. drone micro uav},
  author={Bristeau, Pierre-Jean and Callou, Fran{\c{c}}ois and Vissiere, David and Petit, Nicolas},
  journal={IFAC Proceedings Volumes},
  volume={44},
  number={1},
  pages={1477--1484},
  year={2011},
  publisher={Elsevier}
}

@article{BC09,
  title={Performance of a hopping rotochute},
  author={Beyer, Eric and Costello, Mark},
  journal={Int. Journal of Micro Air Vehicles},
  volume={1},
  number={2},
  pages={121--137},
  year={2009},
  publisher={SAGE Publications Sage UK: London, England}
}

@inproceedings{NT13,
  title={Robust “blind” navigation for a miniature ducted-fan aerial robot},
  author={Naldi \textit{et al}, Roberto},
  booktitle={American Control Conf.},
  pages={988--993},
  organization={IEEE},
  year = {2013}
}

@inproceedings{LL10,
  title={Geometric tracking control of a quadrotor UAV SE (3)},
  author={Lee, Taeyoung and Leok, Melvin and McClamroch, N Harris},
  booktitle={49th IEEE Conf. Decision and Control},
  pages={5420--5425},
  year={2010},
  organization={IEEE}
}

@article{OM15,
  title={A hybrid system framework for unified impedance and admittance control},
  author={Ott, Christian and Mukherjee, Ranjan and Nakamura, Yoshihiko},
  journal={Journal of Intelligent \& Robotic Systems},
  volume={78},
  number={3-4},
  pages={359--375},
  year={2015},
  publisher={Springer}
}

@inproceedings{CD16,
  title={Dynamics of a quadrotor undergoing impact with a wall},
  author={Chui, Fiona and Dicker, Gareth and Sharf, Inna},
  booktitle={Int. Conf. Unmanned Aircraft Systems},
  pages={717--726},
  organization={IEEE},
  year = {2016 }
}

@inproceedings{DC18,
  title={Recovery Control for Quadrotor UAV Colliding with a Pole},
  author={Dicker, Gareth and Sharf, Inna and Rustagi, Pulkit},
  booktitle={Int. Conf. Intelligent Robots and Systems},
  pages={6247--6254},
  organization={IEEE},
  year = {2018 }
}
	
\end{document}